\PassOptionsToPackage{numbers,sort&compress}{natbib}

\documentclass[arxiv]{fairmeta}

\usepackage{xspace}
\makeatletter
\DeclareRobustCommand\onedot{\futurelet\@let@token\@onedot}
\def\@onedot{\ifx\@let@token.\else.\null\fi\xspace}

\makeatother

\usepackage{pifont}
\usepackage{makecell}
\usepackage{array}
\usepackage{amsmath}
\usepackage{amssymb}

\newcommand{\keywords}[1]{}

\newif\ifdraft
\draftfalse

\definecolor{burntorange}{rgb}{0.8, 0.33, 0.0}
\definecolor{orange}{rgb}{1,0.5,0}
\definecolor{green0}{rgb}{0.1,0.7,0.1}

\ifdraft
\newcommand{\PF}[1]{{\color{red}{\bf PF: #1}}}

\newcommand{\CD}[1]{{\color{blue}{\bf CD: #1}}}
\newcommand{\cd}[1]{{\color{blue} #1}}
\newcommand{\AF}[1]{{\color{green0}{\bf AF: #1}}}

\newcommand{\DC}[1]{{\color{orange}{\bf DC: #1}}}

\newcommand{\JK}[1]{{\color{brown}{\bf JK: #1}}}

\newcommand{\TODO}[1]{\textbf{\color{red}[TODO: #1]}}
\else
\newcommand{\PF}[1]{{\color{red}{}}}

\newcommand{\CD}[1]{{\color{blue}{}}}
\newcommand{\DC}[1]{{\color{blue}{}}}
\newcommand{\cd}[1]{ #1 }
\newcommand{\AF}[1]{{\color{green0}{}}}

\newcommand{\JK}[1]{{\color{brown}{}}} 
          
 \newcommand{\TODO}[1]{}
 
\fi

\newcommand{\acron}{GazePrior} 

\newcommand{\parag}[1]{%
  \vspace{0.5em}\noindent\textbf{#1}
}

\newcommand{\truefunc}[0]{{\mathrm{\Psi}}}
\newcommand{\learnedfunc}[0]{\truefunc_\Theta}

\newif\ifcolorticks
\colortickstrue 

\usepackage{xcolor}
\usepackage{amssymb}

\newcommand{\cmark}{%
  \ifcolorticks
    {\color{darkgreen}\checkmark}
  \else
    \checkmark
  \fi
}
\newcommand{\xmark}{%
  \ifcolorticks
    {\color{darkred}\ding{55}}
  \else
    \ding{55}
  \fi
}

\definecolor{darkgreen}{RGB}{0,100,0}
\definecolor{darkred}{RGB}{140,0,0}

\title{GazePrior: Zero-Shot AR/VR Eye Tracking via Learned 3D Gaze Reconstruction}

\author[1\star,2]{Corentin Dumery}
\author[1]{David Colmenares}
\author[1]{Alexander Fix}
\author[2]{Pascal Fua}
\author[1]{Ali Behrooz}
\author[1]{Jogendra Kundu}

\affiliation[1]{Meta Reality Labs, Redmond, USA}
\affiliation[2]{EPFL, Lausanne, Switzerland}

\contribution[\star]{Work done during an internship at Meta}

\abstract{%
Eye tracking (ET) is a foundational technology for advanced AR/VR applications. However, training ET models for every new device is challenging: real data collection is costly and time-consuming, while existing synthetic data generation methods lack realism.
To remove the need for additional data collection while maintaining data quality, we introduce a data-driven 3D prior that models the distribution of human eyes across diverse identities, gaze directions, and light settings.
This model, which we coin \textit{\acron}, then enables sparse-input 3D reconstruction of annotated data collected with \textit{previous} ET devices, which can in turn be rendered from the cameras of any \textit{target} ET device.
Our approach synthesizes data with the realism, diversity, and ground-truth accuracy of real data collection without its prohibitive costs. Our experiments demonstrate that ET models trained with our synthesized data outperform previous zero-shot methods, achieving higher accuracy and robustness.
More information is available on the project page: \url{https://corentindumery.github.io/projects/gazeprior.html}.
}

\metadata[Keywords]{Eye Tracking, Data Synthesis, 3D Reconstruction, Shape Prior, Sparse-Input Reconstruction}

\correspondence{Jogendra Kundu at \email{jogendrak@meta.com}}

\begin{document}

\maketitle


\section{Introduction} \label{sec:intro}


Eye tracking is key to the next-generation of smart glasses and AR/VR headsets. Typical devices~\cite{htcviveproeye,metaquestpro,hololens2,applevisionpro,engel2023aria} rely on up to 5 cameras per eye to reliably and accurately estimate gaze direction and 3D pupil location, with a required precision of the order of a few degrees to enable essential features such as foveated rendering and gaze-based interaction. The captured images are fed to deep networks that have to be properly trained. In theory, this is a simple matter of collecting appropriate training sets. However, in industrial practice, this is an expensive and time-consuming affair because the hardware configuration  of the devices, including camera placement and image resolution, evolve over time, requiring the acquisition of new training sets comprising millions of frames with high-quality ground truth for each new device generation. 

In this paper, we propose leveraging the existence of large-scale datasets with ground-truth annotations that were collected while developing earlier headsets.
\cd{Our goal is to repurpose this data to} train eye tracking (ET) networks that can handle the images acquired by newer headsets featuring different camera setups without having to acquire {\it any} images using the new device, in other words, a  \textit{zero-shot} approach to eye tracking in AR/VR.
To this end, we propose a {\it semi-synthetic} approach to generating training images for the new devices given only the ones acquired using existing older configurations and the new camera specification and configuration. This means that networks can be trained for new devices even before they are built, which will enable precise evaluation of how camera placement will impact ET performance. Thus, it holds the promise of accelerating development without incurring the prohibitive costs associated with large-scale data collection.

\begin{figure}[t]
\centering
\includegraphics[width=\columnwidth]{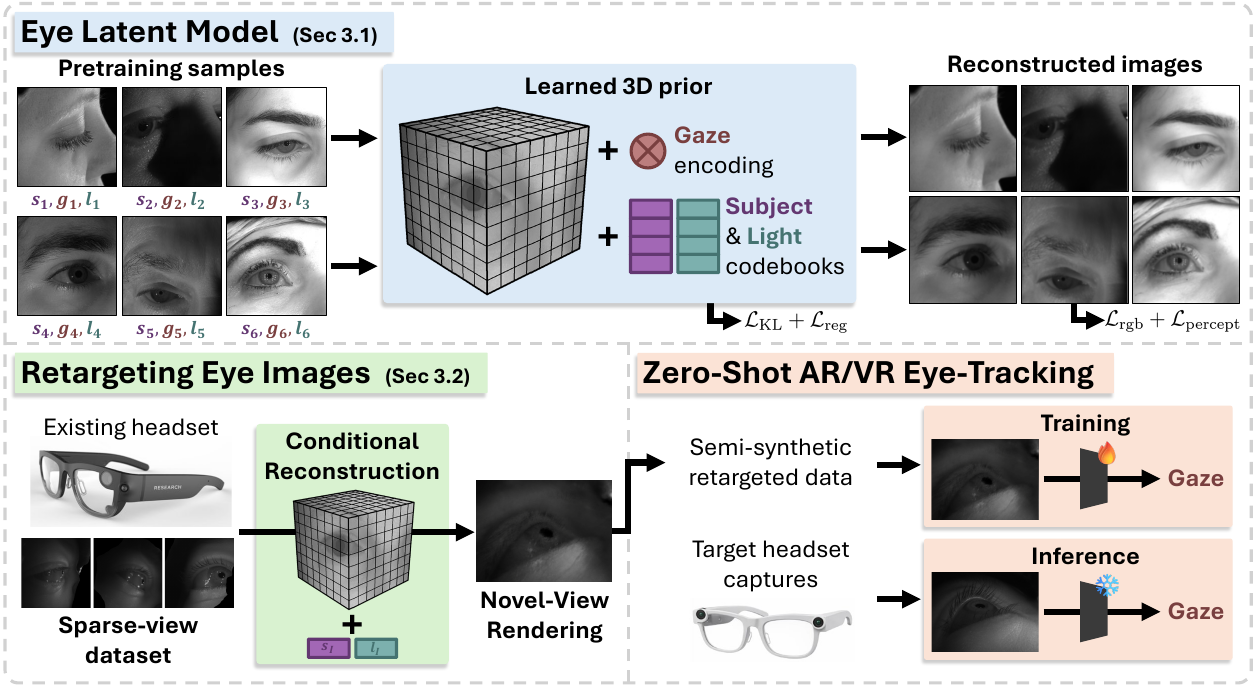} 
\caption{\textbf{Overview of~\acron.} We learn a 3D eye prior (\textit{top}) to achieve high-fidelity retargeting of annotated eye data (\textit{bottom left}). This enables robust eye tracking on new devices (\textit{bottom right}) without requiring any device-specific data collection.}
\label{fig:teaser}
\end{figure}

At the heart of our approach is the following insight. Even though capturing different people using various camera configurations results in a great diversity of images, there are nevertheless strong similarities across subjects, gaze directions, and lighting environments.
Inspired by recent 3D avatar approaches~\cite{buhler2023preface,buehler2024cafca},
we model this prior as a conditional radiance field, which ensures that the generated views maintain spatial and photometric consistency across varying viewpoints.
These methods achieve high-quality avatar reconstruction from only sparse input views by learning a multi-subject latent-vector representation and leveraging it to condition 3D reconstruction.
In this work, we extend this latent-vector approach by conditioning on identity, gaze, and illumination in higher-dimensional latent spaces, and use it to synthesize new training images for a new setup from existing ones. We opt for conditional neural radiance fields for this purpose because they demonstrate superior accuracy across recent benchmarks~\cite{kulhanek2024nerfbaselines} and are well-suited for conditioning on high-dimensional latent embeddings. 


This contrasts with earlier semi-synthetic approaches to eye image synthesis~\cite{ruzzi2023gazenerf,li2025we,lin2025digitally} that reconstruct each new eye from scratch. 
Such methods require dense input views unavailable in standard ET setups, drastically reducing the diversity and quantity of data they can use. 
To compensate, they often synthesize new gaze directions for the few available subjects. 
A typical approach is to rely on explicit eyeball masks to reconstruct the eyeball separately from the face~\cite{wei2025gazegaussian,li2025we,ruzzi2023gazenerf,lin2025digitally}, then rotate the mesh to simulate new gaze directions.  
However, this ignores the intricate coupling between eyeballs, eyelids, and other facial features essential for physically-plausible synthesis, degrading accuracy and limiting usefulness when even small gaze errors disrupt immersion.
Instead, we argue that this coupling should be learned by a network.

In short, state-of-the-art methods for synthesizing eye tracking images have severe limitations that are summarized in Tab.~\ref{tab:comparison_part1_transposed} and which  our approach removes. Our main contributions are summarized as follows:
\begin{itemize}
    \item We are the first to introduce a learned gaze prior that disentangles subject, gaze direction, and light, with biologically accurate eyelid movements. We do not rely on eyeball rotations that limit the realism of synthesized data. 
    \item We incorporate this model into a novel-view-synthesis pipeline that synthesizes images that can be used to train an eye tracker to perform accurately and reliably on real-images. 
    \item We demonstrate the applicability of our synthesized data to zero-shot eye tracking, achieving precise gaze prediction and unlocking new capabilities for models without any device-specific data collection. 
\end{itemize}
To compare our approach for eye tracking purposes against very recent eye-image synthesis baselines~\cite{lin2025digitally,li2025we,wei2025gazegaussian}, we rely on the state-of-the-art ET architecture and system of the Aria~\cite{engel2023aria} headset.
When trained on our images instead of the baselines, it consistently achieves superior performance.

\section{Related Works}
\label{sec:rw}

\parag{Gaze Estimation.}
Estimating 3D gaze direction from \textit{in-the-wild} images has been a long-standing computer vision challenge. Early approaches typically relied on geometric features, employing additional cues like face landmarks~\cite{park2018learning, liu20203d, lu2022neural, zhang2015appearance}, RGB-D images~\cite{mora2013person, funes2016gaze, funes2014eyediap}, or explicitly fitting a morphable 3D eye model to the image data~\cite{wood20163d, wood2018gazedirector, ververas20243dgazenet, kuang2022towards, wen2020accurate}.
The field saw a significant shift with the advent of machine learning and deep learning techniques~\cite{lu2014learning, lu2015gaze}, moving beyond initial linear regression models~\cite{lu2011inferring}. This transition, alongside the creation of the first evaluation benchmarks~\cite{funes2014eyediap,zhang2020eth,zhang2017mpiigaze}, highlighted the critical importance of large, high-quality training data. Several strategies emerged to address this data scarcity: crowdsourcing eye tracking data~\cite{krafka2016eye}, augmenting real datasets by generating novel view angles while preserving the original gaze direction~\cite{ververas20243dgazenet}, or creating vast synthetic datasets using rendering engines to provide precise ground-truth~\cite{wood2015rendering, wood2016learning,bao2025gazegene}. However, crowdsourced real data lacks accuracy, while synthetic data lacks diversity and realism, hindering the performance of downstream eye tracking models in practical scenarios. More recent work went further by attempting to infer gaze when the eyes are not visible, with GA3CE~\cite{kawana2025ga3ce} leveraging 2D segmentations to identify likely 3D gaze targets, and ST-WSGE~\cite{vuillecard2025enhancing} employing pseudo-labeling as a form of weak supervision.

Gaze tracking for Augmented and Virtual Reality (AR/VR)~\cite{clay2019eye, plopski2022eye} is even more challenging. In this context, cameras are at a much smaller distance from the eye and the required precision is significantly higher. Robust gaze tracking is necessary to allow new features in wearable headsets such as foveated rendering~\cite{patney2016towards} and gaze-pinch interaction~\cite{pfeuffer2017gaze, mutasim2021pinch}. The images captured by these eye tracking cameras are rich in context, as demonstrated by egoPPG~\cite{braun2025egoppg} which predicts heartrate from ET images alone. Despite these important applications, AR/VR eye tracking remains comparatively under-explored~\cite{hong2018challenges, david2023privacy}, with some early works using event cameras~\cite{li2023track} or polarized images~\cite{vzurauskas2025polarization}. Our approach is designed to be agnostic to the final eye tracking model or target application, focusing instead on solving the foundational problem of generating large volumes of high-quality training data for new extended reality products without the prohibitive cost of physical data collection.

\parag{Novel View Synthesis.} 
NVS methods that generate unseen views from a set of input images have advanced significantly in recent years and hold the potential to transform eye tracking by generating high-quality eye tracking data at arbitrary viewing angles. The foundational Neural Radiance Fields (NeRFs)~\cite{mildenhall2021nerf} represent a scene in terms of a function that predicts density and color values at every point of a volume and for all possible viewing directions. 
%
%
%
Subsequent works have substantially accelerated both training and inference while simultaneously enhancing reconstruction quality. Key developments include Instant-NGP~\cite{muller2022instant} and 3D Gaussian Splatting (3DGS)~\cite{kerbl20233d}, alongside methods like Mip-NeRF 360~\cite{barron2022mip} and Zip-NeRF~\cite{barron2023zip} that prioritize fidelity over speed. Although 3DGS-based approaches are generally faster, high-fidelity models such as Zip-NeRF still demonstrate superior reconstruction quality over leading 3DGS-based methods like 3DGS-MCMC~\cite{kheradmand20243d}, 3DGUT~\cite{wu20253dgut}, and Scaffold-GS~\cite{lu2024scaffold} across standard benchmarks~\cite{kulhanek2024nerfbaselines}. 3DGS-based methods also suffer from a non-negligible memory footprint, as each Gaussian particle needs to be stored on disk~\cite{niedermayr2024compressed,javed2024temporally}. Neural radiance fields also provide several practical advantages over explicit representations, depending on the application. These include enabling adaptive sampling~\cite{fan2025vcs,li2023nerfacc}, 3D feature fields~\cite{dumery25enforcing,kerr2023lerf}, and model conditioning as demonstrated in our work. Since rendering speed is a secondary concern when generating high-quality training data, we opt for a neural radiance field with state-of-the-art quality improvements~\cite{muller2022instant,buhler2023preface,buehler2024cafca,barron2022mip}.
Finally, we note that multi-view diffusion models~\cite{watsonnovel,chan2023generative} offer a separate path to novel view synthesis, but they often suffer from identity drift~\cite{pooledreamfusion,liu2023zero} and introduce significant alterations to the initial gaze direction due to their lack of underlying 3D representation, rendering them impractical for generating data with accurate ground-truth labels.

\begin{table}[t]
    \centering
    \footnotesize
    \setlength{\tabcolsep}{4pt}
    \begin{tabular}{lcccc}
        \toprule
        \textbf{ } & 
        \makecell{\textbf{Controllable}\\\textbf{Gaze}~\cite{li2025we}} & 
        \makecell{\textbf{Gaze} \\ \textbf{Gaussian}~\cite{wei2025gazegaussian}} & 
        \makecell{\textbf{Lin et al} \\~\cite{lin2025digitally}} & 
        \textbf{\makecell{\acron\\(Ours)}} \\
        \midrule
        \textbf{\makecell{Relightable}} & \cmark & \xmark & \xmark & \cmark \\
        \textbf{\makecell{Eye mask free}} & \xmark & \xmark & \cmark & \cmark \\
        \textbf{\makecell{Eyelid movements}} & \xmark & \xmark & \xmark & \cmark \\
        \textbf{\makecell{Fixed-template free}} & \xmark & \cmark & \cmark & \cmark \\
        \textbf{\makecell{\# per-subject images}} & 1498 & 9849 & 17 & 2-6 \\ 
        \bottomrule
    \end{tabular}
    \vspace{0.5em}
    \caption{Summary of the main key advantages of \acron~over competing methods.}
    \label{tab:comparison_part1_transposed}
\end{table}

\parag{Eye Image Synthesis.} 
Many works attempt to synthesize novel eye images with different gaze directions, which can then be used to generate training data for eye tracking models. Some approaches perform gaze redirection in 2D image space directly, either employing an explicit eye model~\cite{wood2018gazedirector} or utilizing 2D generative networks~\cite{zheng2020self, park2019few, he2019photo, wang2018hierarchical, xia2020controllable}. However, without explicit 3D representations, these models often introduce artifacts or alter the ground truth geometry, rendering them unreliable for XR eye tracking. In a related line of work, earlier methods focused on warping the image to enforce a target gaze direction, such as looking at the camera~\cite{criminisi2003gaze, giger2014gaze, kuster2012gaze, kononenko2015learning} or towards any arbitrary direction~\cite{shu2016eyeopener, ganin2016deepwarp}.

The progress in 3D reconstruction has enabled successful works in high-quality face~\cite{li2017learning, blanz2023morphable, buhler2023preface, ploumpis2019combining, wang2025gaussianhead, xiang2024flashavatar, teotia2024hq3davatar, saito2024relightable,hong2022headnerf,xu2024gaussian} and eye~\cite{berard2014high, kerbiriou2022detailed} capture. Nevertheless, generating large-scale datasets with highly accurate gaze directions using purely real 3D capture remains prohibitively expensive and time-consuming. Instead, some works rely on entirely synthetic datasets~\cite{wood2015rendering, wood2016learning, bao2025gazegene}. More recently, the focus has shifted to generating semi-synthetic training data where samples are synthesized from real images using NVS techniques~\cite{wang2023high, li2022eyenerf, li2024shellnerf, wei2025gazegaussian, li2025we, lin2025digitally, ruzzi2023gazenerf, yin2024nerf}. This approach reduces the domain gap between real and synthetic data, lessening the need for domain adaptation methods~\cite{peng2018synthetic, farahani2021brief}. Notably, EyeNeRF~\cite{li2022eyenerf} introduced a hybrid model that reconstructs the eyeball separately from the rest of the head. This representation was leveraged by GazeNeRF~\cite{ruzzi2023gazenerf} to perform gaze redirection and train an eye tracking network, and later extended by GazeGaussian~\cite{wei2025gazegaussian} and ControllableGaze~\cite{li2025we} to leverage the speed improvements of 3DGS~\cite{kerbl20233d}.

However, these methods require explicit face and eye masks to model the components separately and rotate only the eyeball to generate new ET training samples. This is particularly detrimental because accurate ET predictions rely not only on the eyeball but also on the subtle, correlated movements of the eyelid, eyelashes, and surrounding facial regions, which are ignored by these techniques. Furthermore, most of these algorithms train a separate model per subject, leading to high costs and poor robustness. In contrast, we learn a general latent space that disentangles identity from gaze dynamics and illumination without an explicit representation. This improves the robustness of the NVS, enabling high-quality 3D reconstruction even from sparse ET views.


\section{Method}
\label{sec:method}

Although changes in subjects' appearance, lighting conditions, and camera pose introduce significant diversity in eye images, they still  exhibit commonalities. To exploit these commonalities when synthesizing new images, we take our inspiration from 3D avatar methods~\cite{buhler2023preface,buehler2024cafca} and train a radiance field conditioned on a latent vector that is a high-dimensional embedding of subject identity,  illumination conditions, and gaze orientation, as depicted by \cref{fig:stages}(a). This ensures that the generated views maintain spatial and photometric consistency across varying viewpoints. As shown in \cref{fig:stages}(b,c), the radiance field can then be used to synthesize new images for specific imaging conditions corresponding to different setups, which can in turn be used to train the eye tracker for new devices without having to acquire new images.  

As discussed in the related works section, this is a departure from earlier eye-synthesis methods that either train a separate radiance field per subject~\cite{lin2025digitally,li2025we,ruzzi2023gazenerf,li2022eyenerf}, which requires many images per subject, or rely on separate reconstructions for the eye and face~\cite{li2025we,wei2025gazegaussian,ruzzi2023gazenerf} to rotate the eyeball independently, which neglects that they are correlated and lacks realism.   
In the remainder of this section, we first present our latent-vector model and then discuss how we use it to synthesize new training images. 

\subsection{Eye Latent Model}
\label{sec:latent}

A typical approach to learning radiance fields for image synthesis is to represent them as a neural radiance function 
%
%
\begin{equation}
\label{eq:nerf}
\truefunc_{\Theta} \left(x, v\right) = 
\left( \sigma(x), \, c(x, v) \right)
\end{equation}
that takes as input a 3D point $x$ and view direction $v$, and outputs a density $\sigma$ and color $c$, with $\Theta$ the weights of the network that implement $\truefunc$. Given a set of input images, these weights are optimized so that colors and densities can be used to re-synthesize the images using volume rendering~\cite{mildenhall2021nerf}.  Usually, this is done for several images of the same subject acquired using cameras at different locations, but with everything else being held constant. In~\cite{buhler2023preface,buehler2024cafca}, $\truefunc$ takes an additional argument that represents the identity of a person so that the same set of weights $\Theta$ can be used to synthesize images of different people. In this work, we further extend this to multiple subjects and gazes under changing illumination conditions by making the subject's identity, lighting, and gaze directions input to the network. 

\parag{Neural Radiance Function $\truefunc$.} 
Let us assume that we are given $N$ frames $\{F_i ,s_i, g_i, l_i\}_{i \in [1,N]}$, each featuring a set of images $F_i$ of subject $s_i$, with gaze direction~$g_i$ and light setting~$l_i$. We make $\truefunc$ dependent on these and rewrite Eq.~\ref{eq:nerf} as
%
%
\begin{equation}
\label{eq:ext_nerf}
\truefunc_{\Theta} \left(x, v, \hat{s}_i, \hat{g}_i, \hat{l}_i\right) = 
\left( \sigma(x, \hat{s}_i, \hat{g}_i), \, c(x, v, \hat{s}_i, \hat{g}_i, \hat{l}_i) \right) \; ,
\end{equation}
where $\hat{s}_i$, $\hat{g_i}$, $\hat{l_i}$ are high-dimensional embeddings that encode ${s_i}$, ${g_i}$, $l_i$. They function as latent variables and are discussed in more detail below. 

Note that the formulation of Eq.~\ref{eq:ext_nerf} disentangles the latent variables for identity from gaze direction and illumination. This has several benefits.  First, it enables us to define each embedding independently. Second, it makes our model both interpretable and controllable. Last, as demonstrated by our experimental results, this representation is relatively easy to learn for a multi-subject model, which significantly enhances novel view synthesis performance.

\begin{figure}[t]
\centering
 \includegraphics[width=0.99\linewidth]{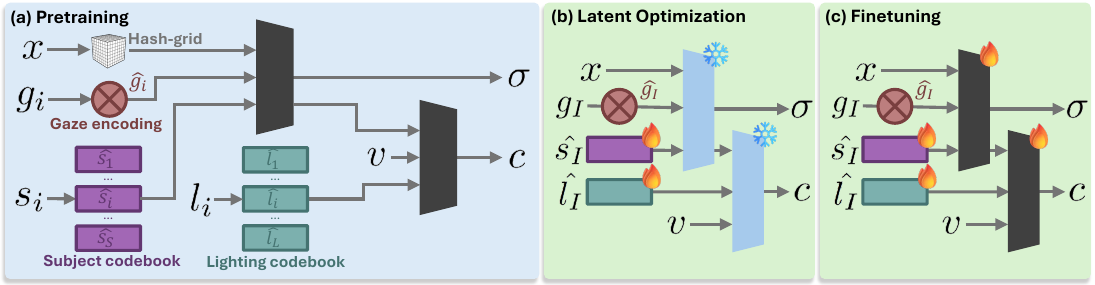} 
 \caption{\textbf{Model.} We learn a 3D prior from multiple identities, gaze directions, and lightings during pretaining (a). To obtain a high-quality reconstruction from a new set of images $I$, we first optimize latent codes (b), then finetune the complete model (c).}
 \label{fig:stages}
\end{figure}

Fig.~\ref{fig:stages}(a) depicts the network that implements $\truefunc$. It comprises a {\it density network} that takes as input $x$, $\hat{g}_i$, and $\hat{s}_i$ to return $\sigma$ along with a second output that is fed to a {\it color network}, along with $\hat{l}_i$ and $v$, to produce $c$. In other words $\hat{l}_i$ does not have any impact on $\sigma$, only on $c$. This architecture has several advantages. First, it provides regularization in darker settings that are notoriously challenging for radiance field optimization but prevalent in real-world eye tracking data \cd{captured under reduced illumination within head-mounted displays}. Second, it enables the density network to focus entirely on capturing geometric variations across subjects and gaze directions, without the confounding influence of shadows or highlights. Finally, it enhances the interpretability of the latent space, and formally guarantees that interpolations in $\hat{l}_i$ only affect appearance. As demonstrated in our ablation study in~\cref{sec:experiments}, this asymmetric treatment of lighting, identity, and gaze direction brings significant performance gains over treating all embeddings similarly.


\parag{Subject Latent Vector $\hat{s}_i$.} 
To handle many subjects, as opposed to a single one as in many previous works, we represent the subject identity with a high-dimensional vector $\hat{s_i}$ stored in a codebook and indexed by $s_i$. This codebook is learned during the pretraining discussed above without enforcing any structure in the resulting latent space to let the model learn the subtle difference and similarities between subjects.


While dealing with hundreds of subjects given a fixed model capacity reduces per-subject reconstruction detail, it imposes a beneficial regularization. To accommodate diverse identities, the network has incentive to learn a shared, low-frequency representation. This serves as a powerful regularizer that effectively reduces the appearance of \textit{floaters} and promotes the recovery of smooth, geometrically consistent surfaces. Furthermore, this inherent smoothness bias ensures that the resulting latent space is well-behaved, facilitating plausible identity interpolations and robust generalization to unseen subjects.

\parag{Gaze Latent Vector $\hat{g}_i$.}
Gaze direction changes are intricately coupled with non-rigid movements of the eyelids, eyebrows, and other subtle facial features. Unlike previous approaches that treat gaze as a purely rigid transformation of the eyeball~\cite{li2022eyenerf,wei2025gazegaussian,li2025we}, we argue that these complex movements must be learned by a network with an underlying 3D representation, which can capture these complexities while staying consistent to a given 3D subject. This becomes particularly important when the eyelids and eyelashes occlude parts of the pupil, as these subtle cues become the primary signals for precise gaze estimation.


Since gaze directions are continuous, unlike subject identities, we do not use a discrete latent codebook to encode them. Instead, we employ sinusoidal encoding~\cite{tancik2020fourier} to represent $g_i$ as
\begin{align}
\hat{g}_i = [\sin(f_1 g_i), \cos(f_1 g_i), \ldots, \sin(f_F g_i), \cos(f_F g_i)] \; , \label{eq:gaze}
\end{align}
where $[f_i]_{i=1}^F$ is a fixed set of frequencies. \cd{As opposed to simply using a gaze codebook, this sinusoidal encoding allows us to directly leverage the ground-truth gaze present in previously collected ET data, as discussed in \cref{sec:retargeting}. 


}
 


\parag{Lighting Latent Vector $\hat{l}_i$.} 
\cd{Similarly to the subject latent code $\hat{s}_i$, we initialize a learnable codebook of light latents and index the codebook to retrieve $\hat{l}_i$.} 
However, recall from Eq.~\ref{eq:ext_nerf} that the latent light vector, unlike the subject and gaze latent vectors, is not fed to the density network but directly to the color network, as shown in \cref{fig:stages}(a). Consequently, the reconstructed geometry is guaranteed to be independent from the value of $\hat{l}_i$. 

\parag{Pretraining.}
To learn the weights $\Theta$ of the prior $\learnedfunc$ and the latent embeddings, which we will refer to as {\it pretraining}, we leverage the dataset of \cite{lin2025digitally}. It includes more than 400 subjects with multi-view images and variations in gaze direction and lighting.

In practice, we load 16 subjects at a time and reload a new batch every 2000 iterations. To improve the interpretability of our latent space, we also apply a Kullback–Leibler divergence loss $\mathcal{L}_{\text{KL}}$ which encourages the latent codes to follow a normal distribution. We also apply an L2-regularization $\mathcal{L}_{\text{reg}}$ on the weights of the color network to prevent overfitting. We sample rays across frames, and compute an image loss $\mathcal{L}_{\text{rgb}} = |c - \hat{c}|$ where $c$ is the ground truth pixel color and $\hat{c}$ is the predicted color, along with a perceptual loss $\mathcal{L}_{\text{percept}}$ \cite{zhang2018unreasonable}. The final loss is expressed as
\begin{equation}
\mathcal{L} = \mathcal{L}_{\text{rgb}} + \lambda_{\text{percept}} \mathcal{L}_{\text{percept}} + \lambda_{\text{KL}} \mathcal{L}_{\text{KL}} + \lambda_{\text{reg}} \mathcal{L}_{\text{reg}} \; ,
\end{equation}
where $\lambda_{\text{percept}}$, $\lambda_{\text{KL}}$, and $\lambda_{\text{reg}})$ are scalars. Finally, we activate finer hash-grid~\cite{muller2022instant} resolutions progressively throughout pretraining, which encourages the model to learn a coarse, shared representation.

\subsection{Retargeting Eye Images} 
\label{sec:retargeting}

The central goal of this work is to develop a \textit{retargeting} pipeline that makes it possible to use high-quality annotated data from previously released headsets and to re-render them from the specific camera of a new \textit{target} system while maintaining alignment with ground-truth annotations.  To this end, during the pretraining phase of the previous section, we have learned the weights of $\learnedfunc$, which models the manifold of human eye appearance and geometry while disentangling identity, gaze, and illumination. We now use it to conditionally reconstruct high-fidelity radiance fields from the available sparse views. The re-rendering process involves the following steps.

\parag{Initialization.} 
Given a set of source images $I$ with known camera parameters and ground-truth gaze direction $g_I$, we aim to find a reconstruction $\truefunc_{\Theta, \hat{s_I}, \hat{g_I}, \hat{l_I}}$ that fits the identity, gaze, and appearance of the source and can be rendered from any target viewpoint. 
Since the ground-truth gaze direction $g_I$ is known in ET images, we first compute the gaze embedding $\hat{g_I}$ of Eq.~\ref{eq:gaze}, which will remain fixed during the rest of the computation. Then, we align the 3D prior with the source coordinate system. 
We transform both the prior and the source cameras into a device-centric reference frame called the Central Pupil Frame (CPF), where the origin is defined as the midpoint between the left and right pupils \cd{as present in the dataset or otherwise estimated with FLAME~\cite{li2017learning}}. Finally, we initialize latent codes $(\hat{s_I},\hat{l_I})$ for identity and illumination as the mean vectors of their respective codebooks. They are close to zero because minimizing $\mathcal{L}_{\text{KL}}$ during pretraining encourages a zero-mean.

\begin{figure}[t]
    \centering
    \hfill
    \hspace{0.9em}
    \begin{minipage}[t]{0.22\textwidth}
        \centering
        \textbf{Training Views}\\[0.2em]
        \begin{tabular}{>{\centering\arraybackslash}m{0.45\linewidth} >{\centering\arraybackslash}m{0.45\linewidth}}
            \includegraphics[width=\linewidth]{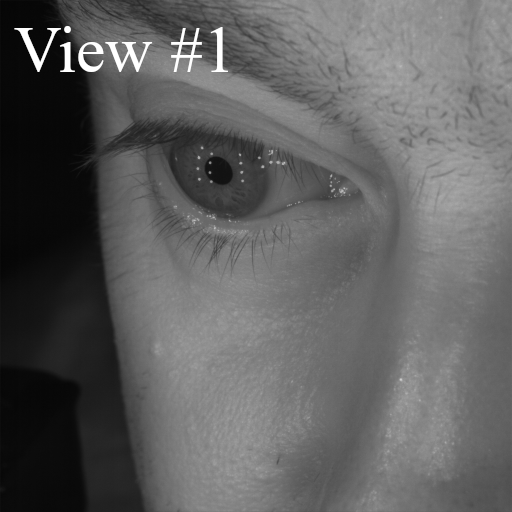} & 
            \includegraphics[width=\linewidth]{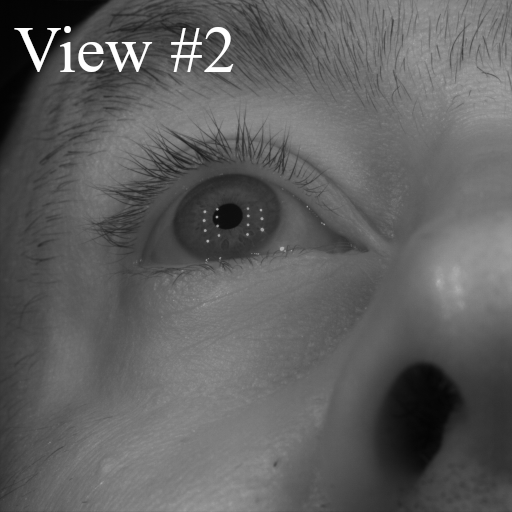} \\
            \includegraphics[width=\linewidth]{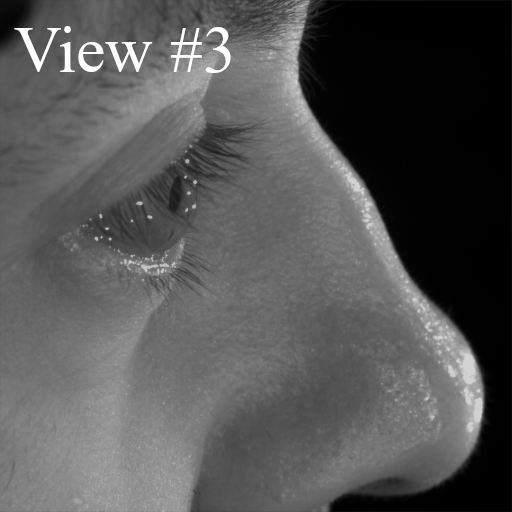} & 
            \includegraphics[width=\linewidth]{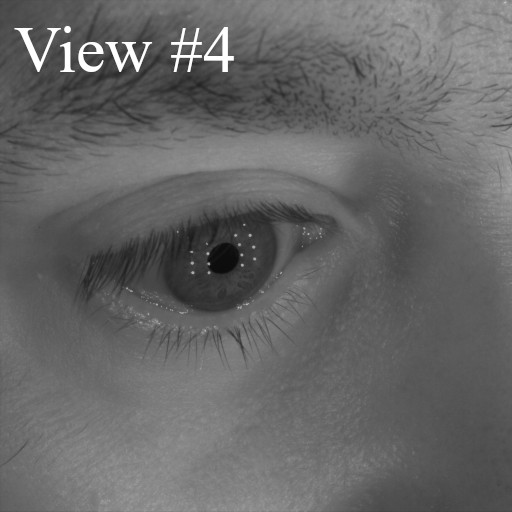} \\
            \includegraphics[width=\linewidth]{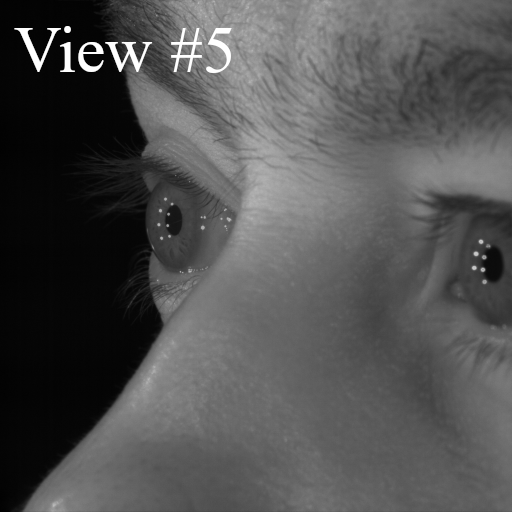} & 
            \dots 
        \end{tabular}
    \end{minipage}%
    \hfill
    \begin{minipage}[t]{0.66\textwidth}
        \centering
        \textbf{Novel-View-Synthesis Performance}\\[0.0em] 
        \includegraphics[width=\textwidth]{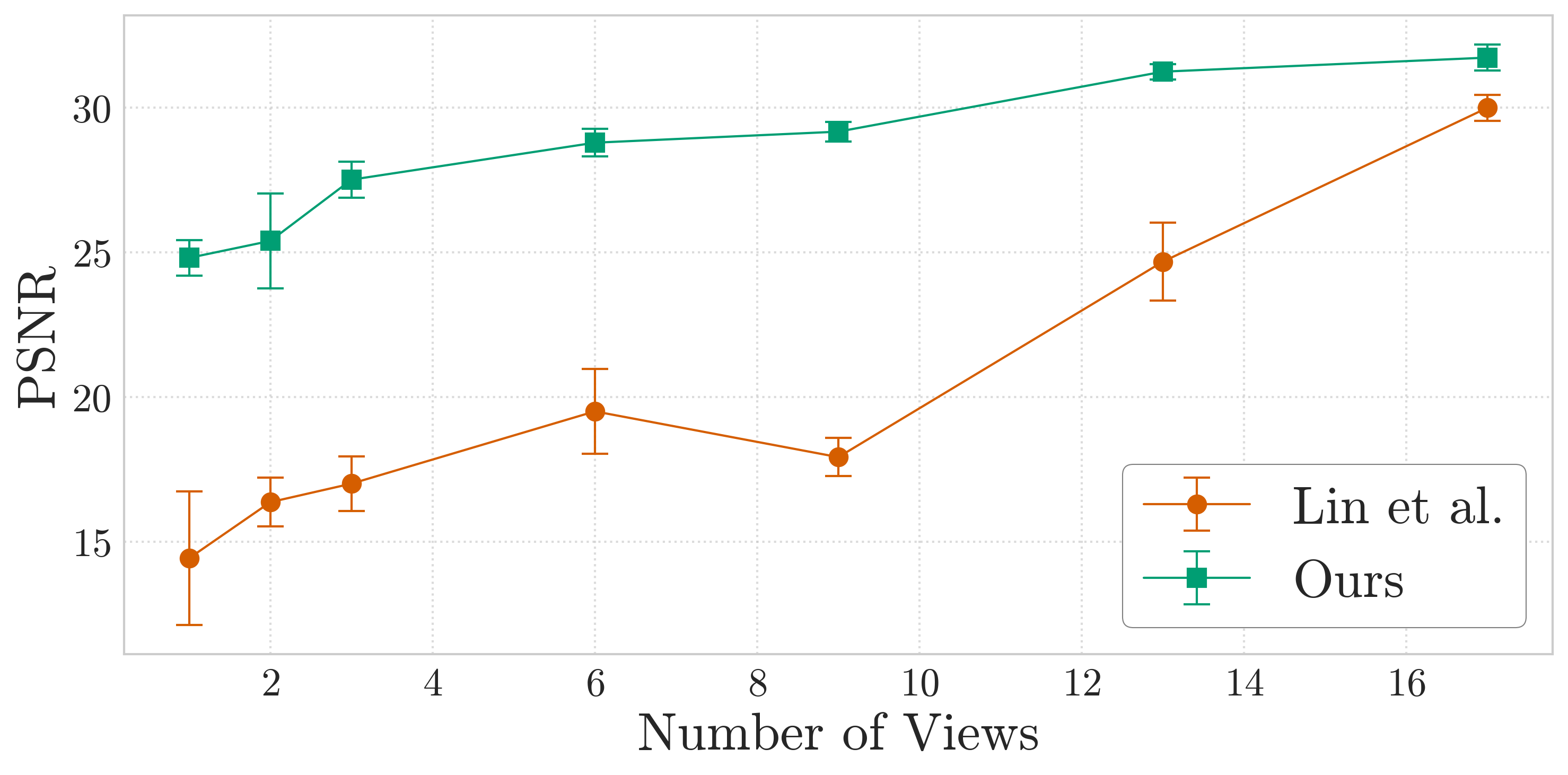}
    \end{minipage}%
    \hfill
    \vspace{0.5em}
    \begin{minipage}[t]{0.98\textwidth}
        \centering
        \begin{tabular}{c@{\hskip 0.5em}c@{\hskip 0.2em}c@{\hskip 0.2em}c@{\hskip 0.2em}c@{\hskip 0.2em}c}
            & \textbf{1 view} & \textbf{2 views} & \textbf{3 views} & \textbf{6 views} & \textbf{17 views} \\
            \rotatebox{90}{\hspace{0.0em}\footnotesize\textbf{Lin et al. \cite{lin2025digitally}}} &
            \includegraphics[width=0.18\textwidth]{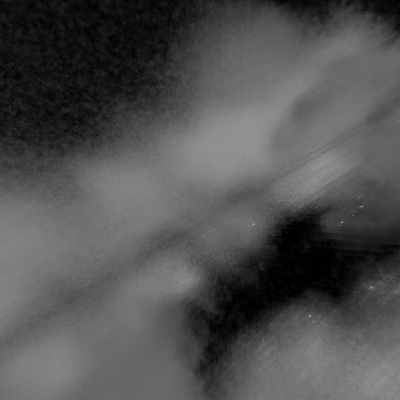} &
            \includegraphics[width=0.18\textwidth]{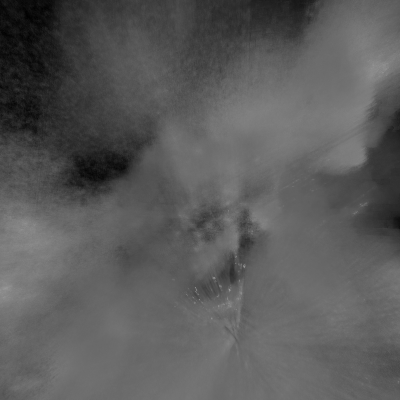} &
            \includegraphics[width=0.18\textwidth]{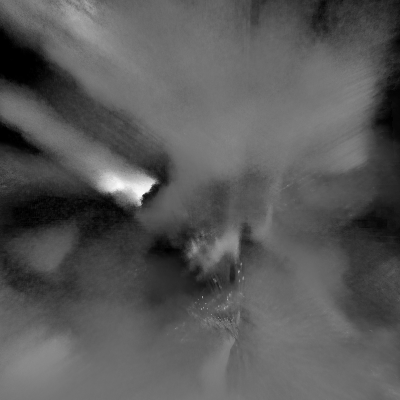} &
            \includegraphics[width=0.18\textwidth]{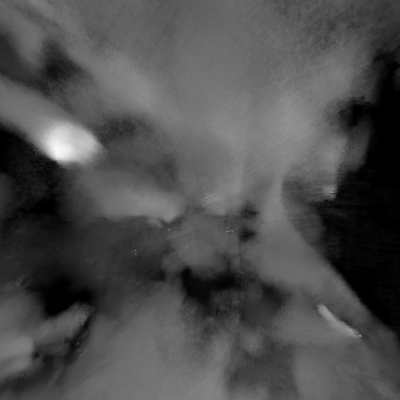} &
            \includegraphics[width=0.18\textwidth]{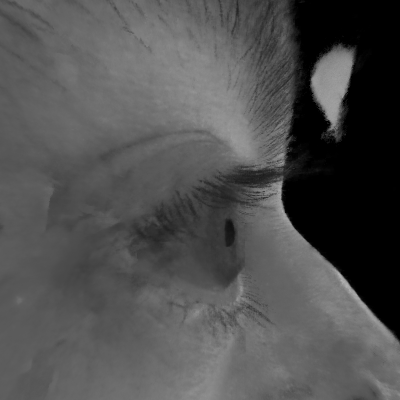} \\
            \rotatebox{90}{\hspace{2em}\footnotesize\textbf{Ours}} &
            \includegraphics[width=0.18\textwidth]{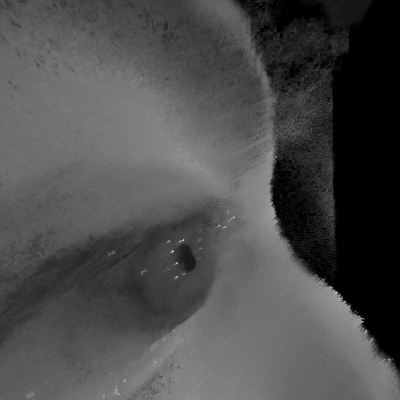} &
            \includegraphics[width=0.18\textwidth]{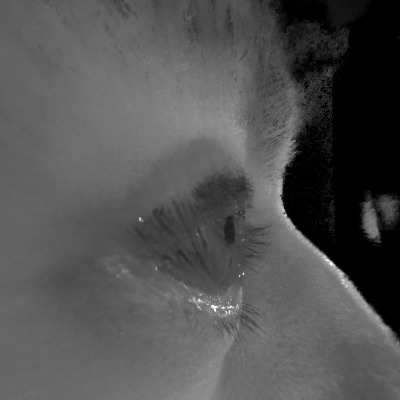} &
            \includegraphics[width=0.18\textwidth]{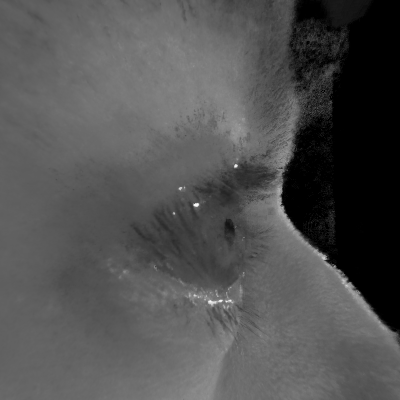} &
            \includegraphics[width=0.18\textwidth]{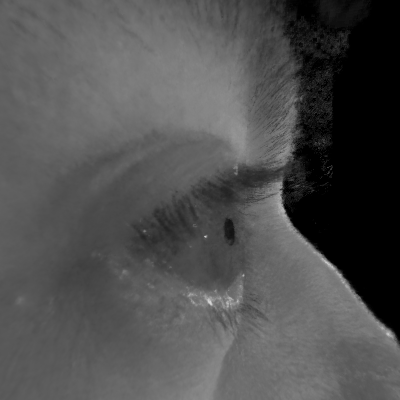} &
            \includegraphics[width=0.18\textwidth]{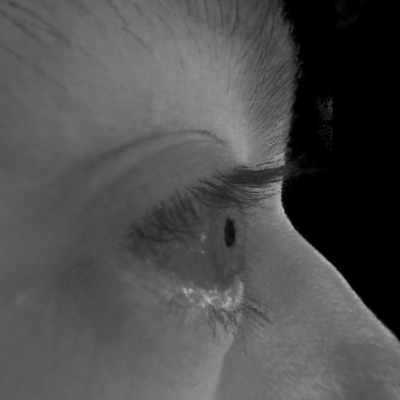} \\
        \end{tabular}
    \end{minipage}
    \vspace{-0.5em}
    \caption{\textbf{View sparsification in finetuning}. Top left: Sparse training views. Top right: Quantitative performance comparison. Bottom: Qualitative comparison between Lin et al.~\cite{lin2025digitally} and our method across varying view counts.}
    \label{fig:view_sparsification}
\end{figure}


\parag{Latent Optimization.}
To maximize the inductive bias of our 3D prior, we start by applying latent optimization with frozen model weights $\learnedfunc$ as in~\cite{buhler2023preface,buehler2024cafca}. We first freeze the weights of $\learnedfunc$ and optimize only the latent codes $(\hat{s_I}, \hat{l_I})$. We sample $2048$ rays across all available source views per iteration and minimize the image reconstruction loss $\mathcal{L}_{\text{rgb}}$. This ensures that the global structure, such as eye shape and lighting environment, is recovered while strictly adhering to the learned 3D manifold and the known ground-truth gaze direction $g_I$. We find that $50$ iterations are sufficient to reach a satisfying starting point without adding a significant computational overhead.

\parag{Finetuning.}
Given the latent codes estimated in this manner, we unfreeze the model to finetune a high-quality radiance field and capture person-specific details that the general prior might omit. During this stage, we jointly optimize the weights of $\truefunc_{\Theta, I}$ and the latent codes $(\hat{s_I}, \hat{l_I})$ using the same loss $\mathcal{L} = \mathcal{L}_{\text{rgb}} + \lambda_{\text{percept}} \mathcal{L}_{\text{percept}} + \lambda_{\text{reg}} \mathcal{L}_{\text{reg}}$ as during pretraining, but without $\mathcal{L}_{\text{KL}}$ since there is a single latent code in this stage. We found experimentally that the pretraining of \cref{sec:latent}  imposes a strong enough inductive bias so that we do not need to enforce any additional constraints on the weights of $\truefunc_{\Theta, I}$ at this stage, which would hinder its ability to fit the reconstruction to the input images.

\parag{Data Synthesis.}
The result is a high-quality 3D radiance field, which we can re-render from the perspective of any target camera while preserving the original ground-truth gaze and pupil annotations. Crucially, the optimization and finetuning stages are performed only once per capture, and the resulting weights $\truefunc_{\Theta, \hat{s_I}, \hat{g_I}, \hat{l_I}}$ can then be used to synthesize a variety of training samples across different viewpoints and camera specifications.

\section{Experiments} \label{sec:experiments}

\textbf{Baselines.}
In our experiments, we use ControllableGaze~\cite{li2025we} and the method of Lin \textit{et al.}~\cite{lin2025digitally} as baselines for both eye image synthesis and eye tracking. These methods are the most recent and have demonstrated their superiority over previous approaches to gaze redirection \cite{ruzzi2023gazenerf,zheng2020self,hong2022headnerf,xu2024gaussian}, synthetic data generation \cite{wood2015rendering,wood2016learning}, generative 2D models \cite{wang2018hierarchical}, and deformable eyes \cite{kuang2022towards,li2024shellnerf}.
However, these baselines still require many views to produce useful reconstructions, limiting the sources of data that they can be trained with. We will show that our method can operate with far fewer, which makes it far more practical. \cd{Additional details regarding how baseline datasets are generated can be found in our supplementary.}


\textbf{Datasets.}
We use the dataset of Lin \textit{et al.} \cite{lin2025digitally} to pretrain \acron. It includes over 400 subjects with variations in gaze direction and light settings. For our \textit{target} ET system, we use the Aria project \cite{engel2023aria}. It relies on a single temporal-view camera oriented directly towards the pupil. \cd{Finally, recall that our method retargets eye tracking images by finetuning the prior model on previous captures from an existing headset. To this end, we leverage a dataset of ET sessions that comprises 5 cameras, with three frame cameras and two distant cameras.
This dataset will be referred to as the retargeting dataset.} We provide additional details in the supplementary.

\subsection{Novel View Synthesis}

\begin{table}[t]
    \centering
    \setlength{\tabcolsep}{3pt} 
    \begin{tabular}{l ccc ccc}
    \toprule
    & \multicolumn{3}{c}{Pretraining} & \multicolumn{3}{c}{Sparse-View Finetuning} \\
    \cmidrule(lr){2-4} \cmidrule(lr){5-7}
    Method & PSNR $\uparrow$ & SSIM $\uparrow$ & MSE $\downarrow$ & PSNR $\uparrow$ & SSIM $\uparrow$ & MSE $\downarrow$ \\
    \midrule
    Lin et al.~\cite{lin2025digitally} & -     & -              & -      & 15.31 & 0.454 & 0.0383 \\
    \textbf{\acron~(Ours)}  & 25.32      & \textbf{0.559} & 0.0044         & \textbf{22.10} & 0.671  & \textbf{0.0064} \\
    \midrule
    w/o disentangled latents      & 18.97 & 0.514          & 0.0166 & 19.09 & 0.647 & 0.0137 \\
    w/o light-independent density & 25.22 & 0.561 & 0.0045 & 21.53 & \textbf{0.682 }& 0.0079 \\
    w/o fixed gaze encoding & \textbf{25.46} & 0.557 & \textbf{0.0043} & 20.00 & 0.679 & 0.0104 \\
    \bottomrule
    \end{tabular}
    \vspace{1mm}
    \caption{\textbf{Novel-view-synthesis evaluation and ablation study.} We serately evaluate pretraining~(\cref{sec:latent}) and finetuning~(\cref{sec:retargeting}) performance.}
    \label{tab:ablation}
\end{table} 

\subsubsection{Quantitative Evaluation.}
\label{sec:quantitative}

We quantify image synthesis performance using decreasing number of views and report the results in~\cref{fig:view_sparsification} in terms of Peak Signal to Noise Ratio (PSNR). When using 17 views, our performance is only slightly better than that of~\cite{lin2025digitally}. However, as the number of views decreases, the gap in performance grows from 1.5 dB to almost 10.0 dB when using only two views. This confirms the importance of our learned prior for sparse-view reconstruction.


In the the three rightmost columns of ~\cref{tab:ablation}, we report comparative results  using 4 views from the retargeting dataset measured in PSNR, structural similarity (SSIM), and Mean Square Error (MSE). We provide the exact definitions of these metrics in our supplementary material. These results are consistent with those of~\cref{fig:view_sparsification}.  We did not include comparisons with ControllableGaze~\cite{li2025we} because it is trained on datasets with several thousands of images per subject and fails when applied on 4 views.

\begin{figure*}[t]
\centering

\setlength{\tabcolsep}{1pt} 

\begin{tabular}{@{} m{0.15\linewidth} m{0.13\linewidth} m{0.13\linewidth} m{0.13\linewidth} m{0.13\linewidth} m{0.13\linewidth} m{0.13\linewidth} @{}}

\centering \footnotesize Controllable Gaze~\cite{li2025we} &
\includegraphics[width=\linewidth]{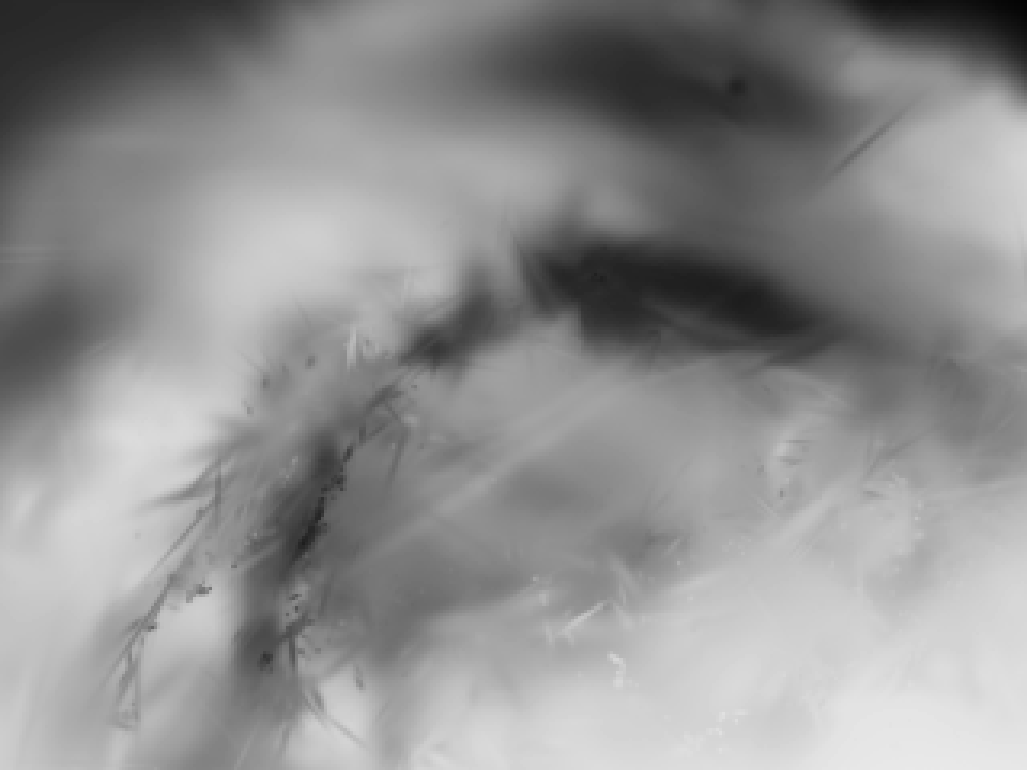} &
\includegraphics[width=\linewidth]{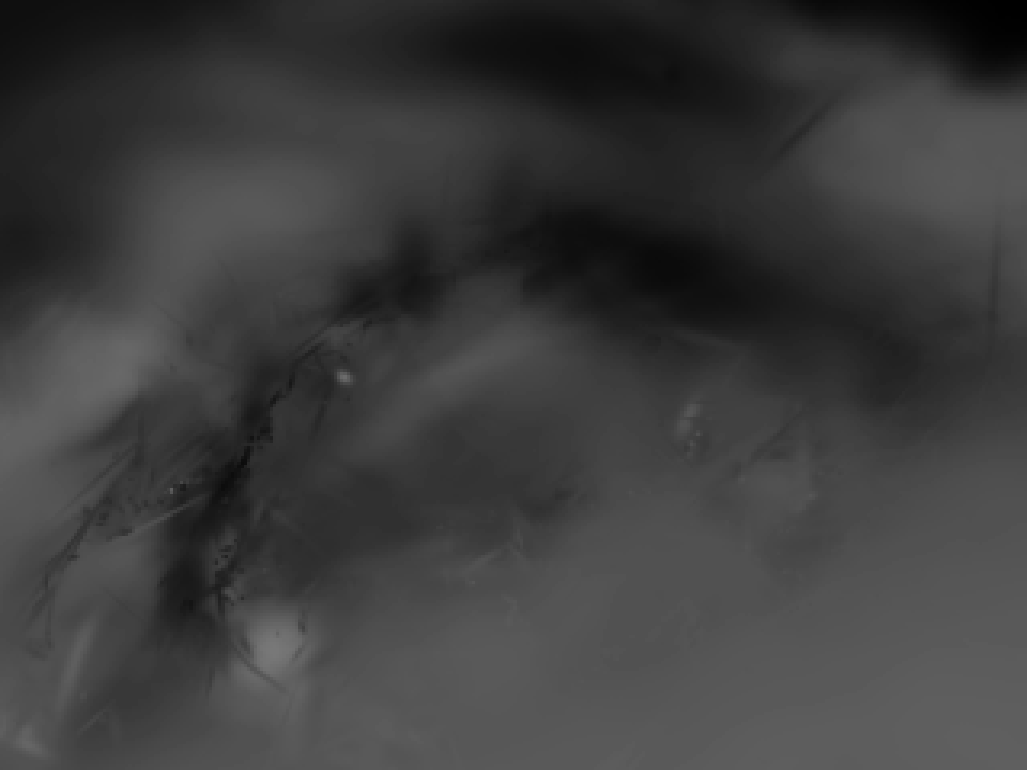} &
\includegraphics[width=\linewidth]{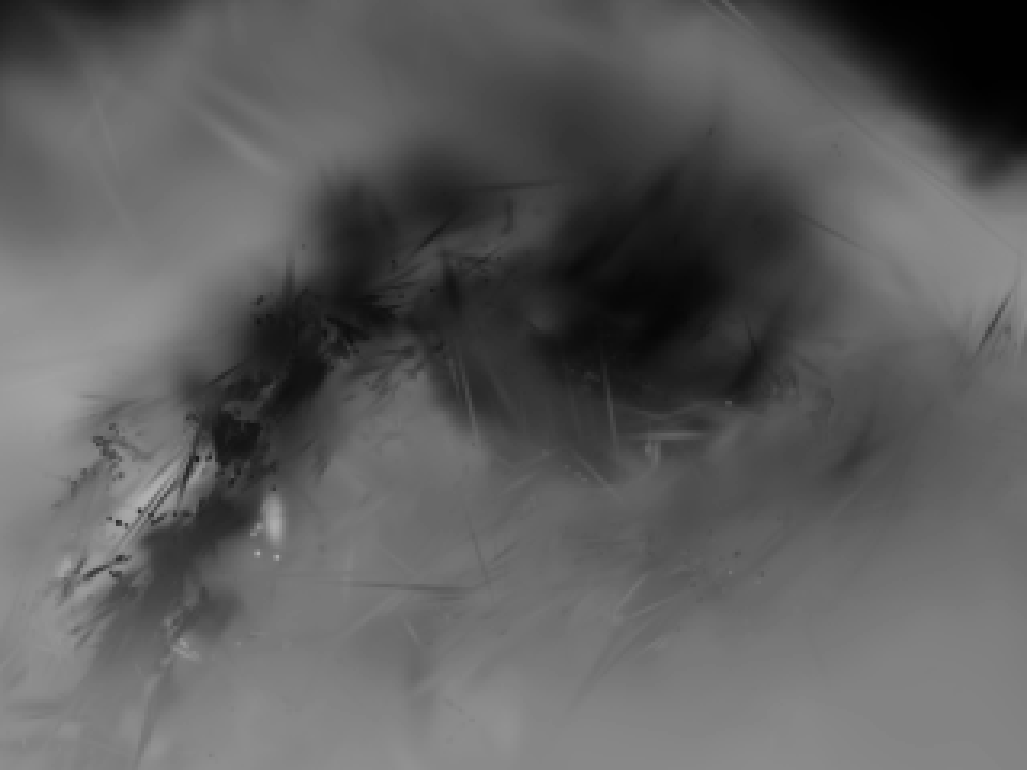} &
\includegraphics[width=\linewidth]{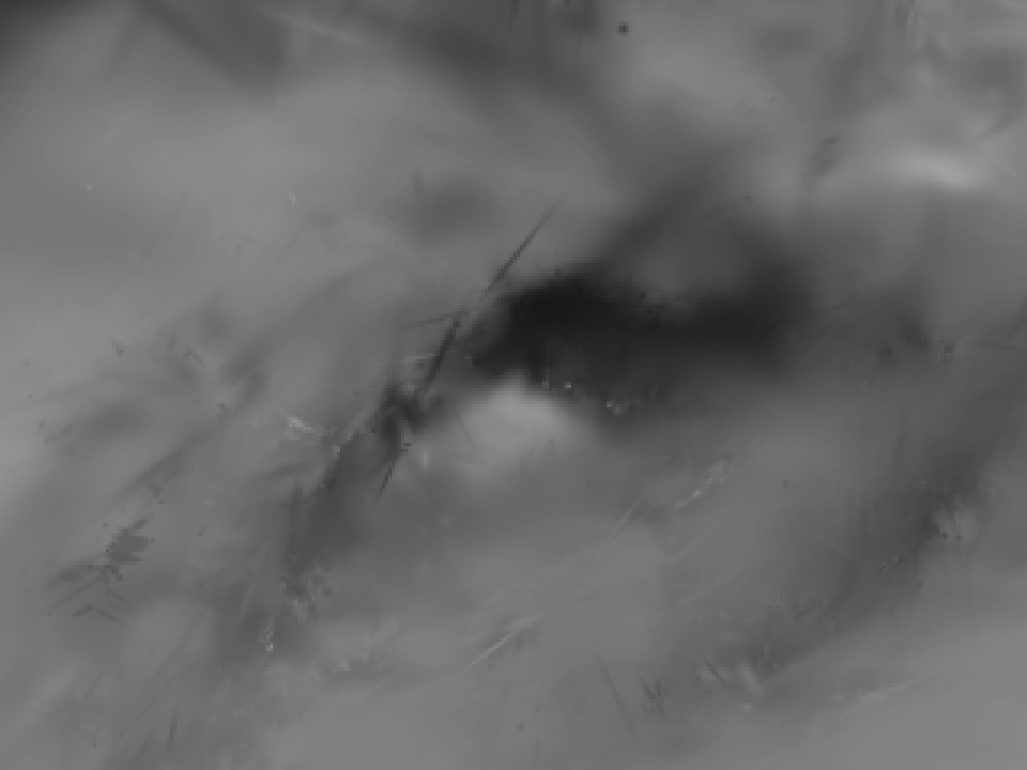} &
\includegraphics[width=\linewidth]{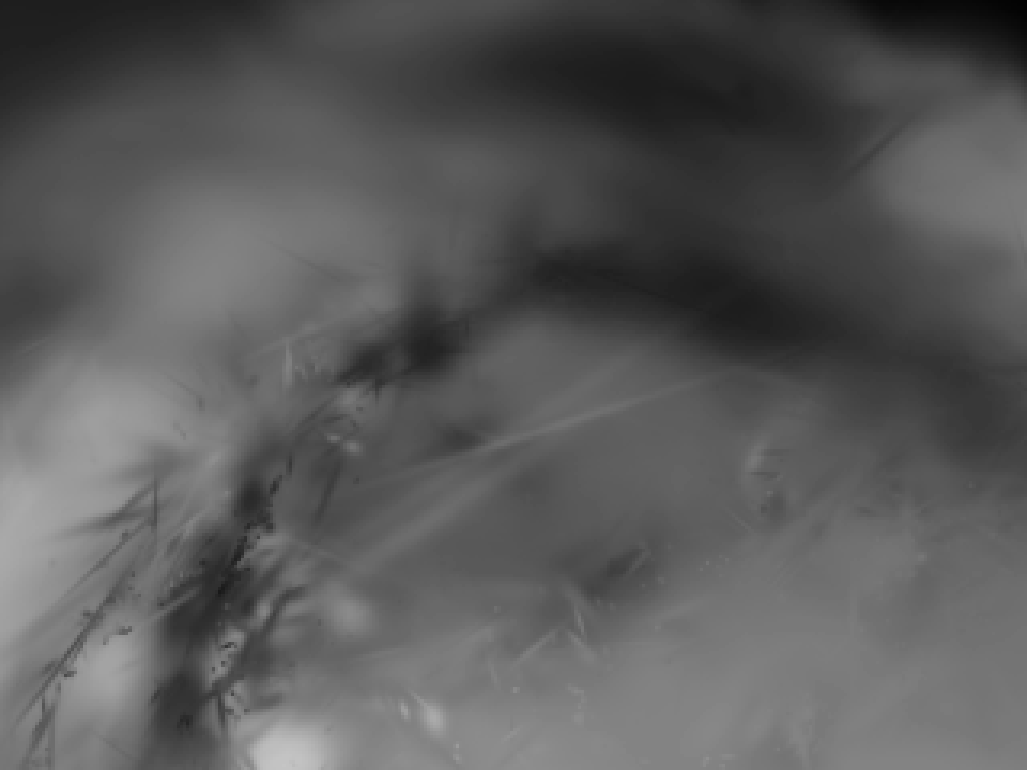} & 
\includegraphics[width=\linewidth]{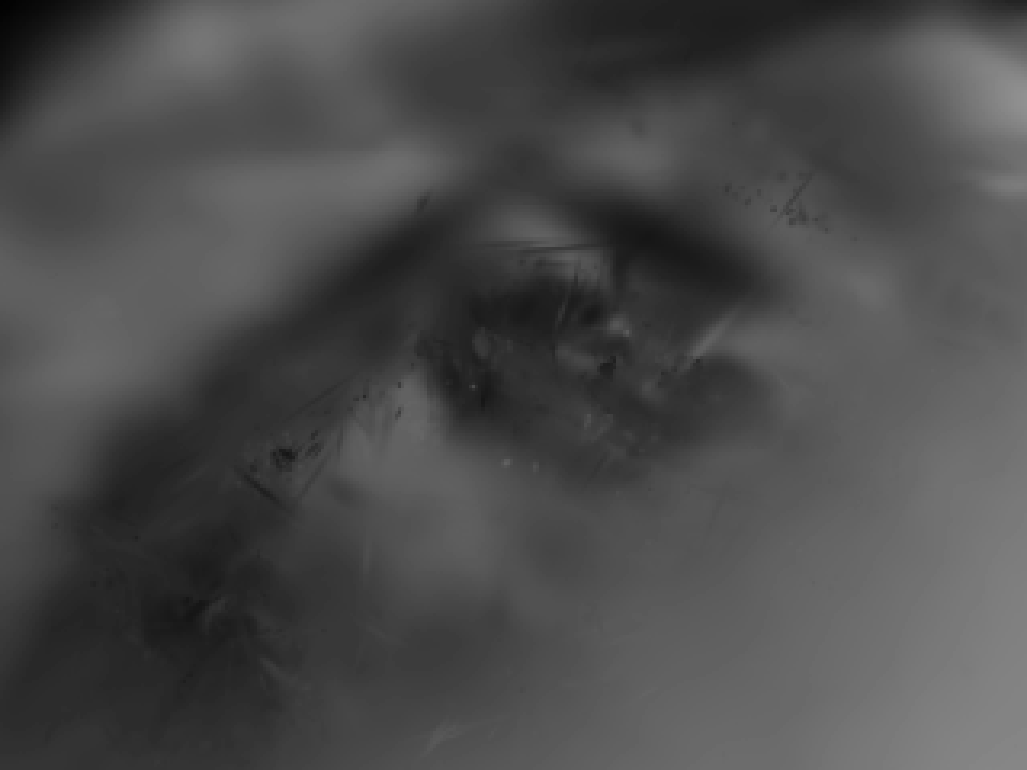} \\

\centering \footnotesize Lin \textit{et al.}\\\cite{lin2025digitally} &
\includegraphics[width=\linewidth]{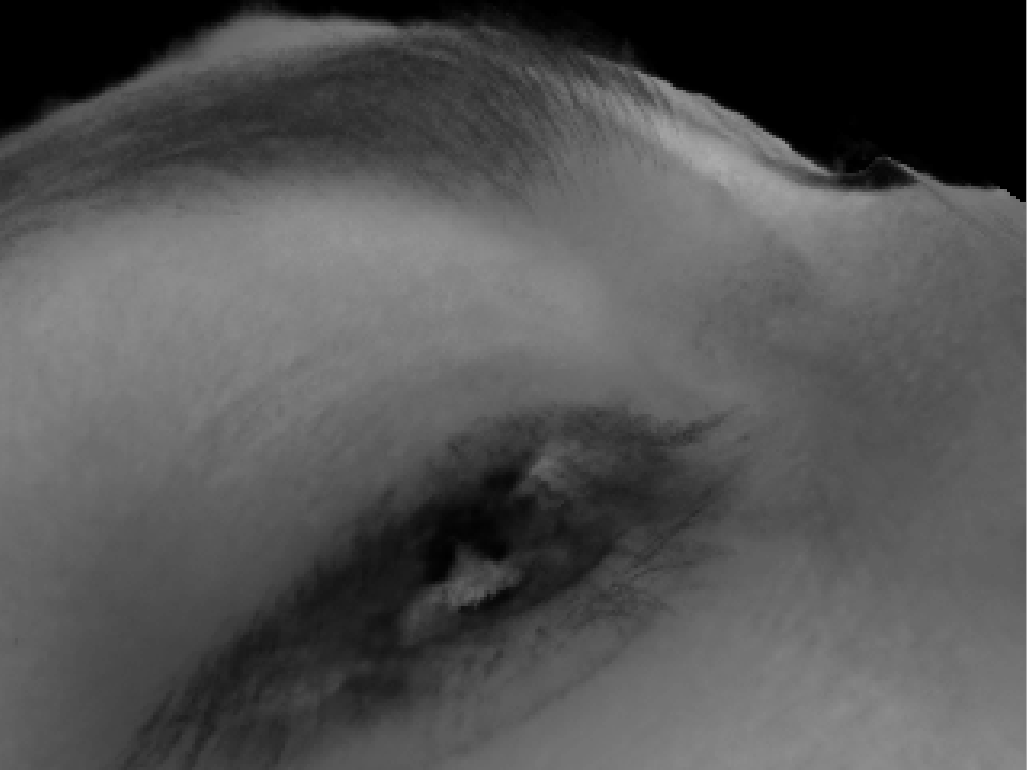} &
\includegraphics[width=\linewidth]{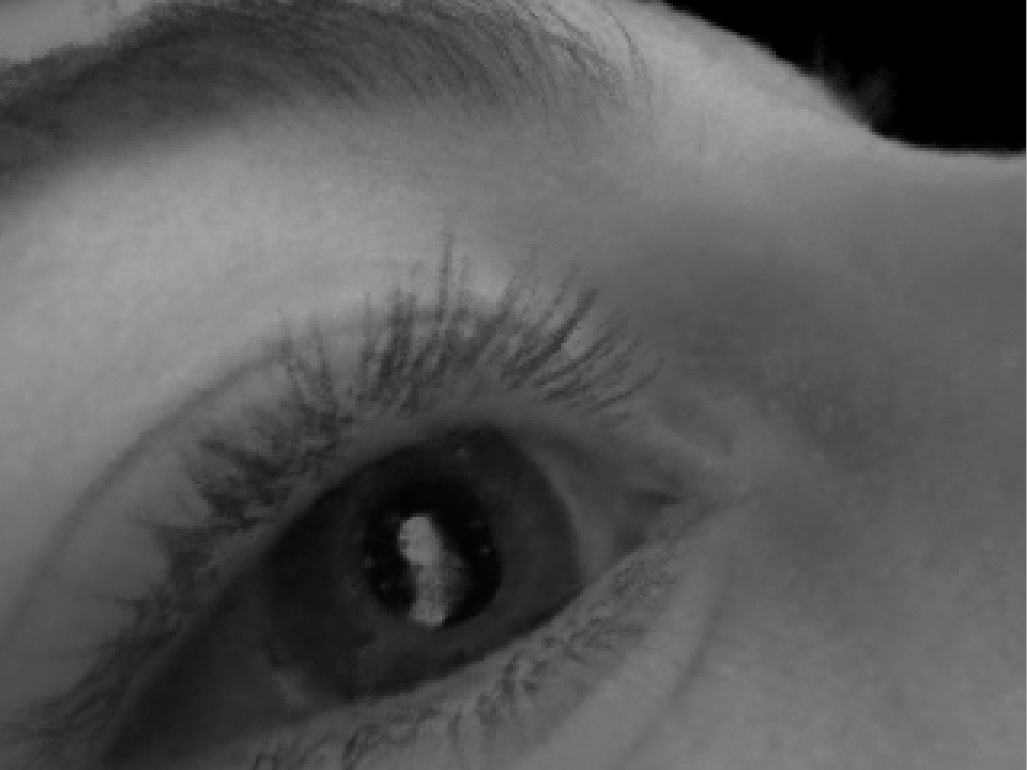} &
\includegraphics[width=\linewidth]{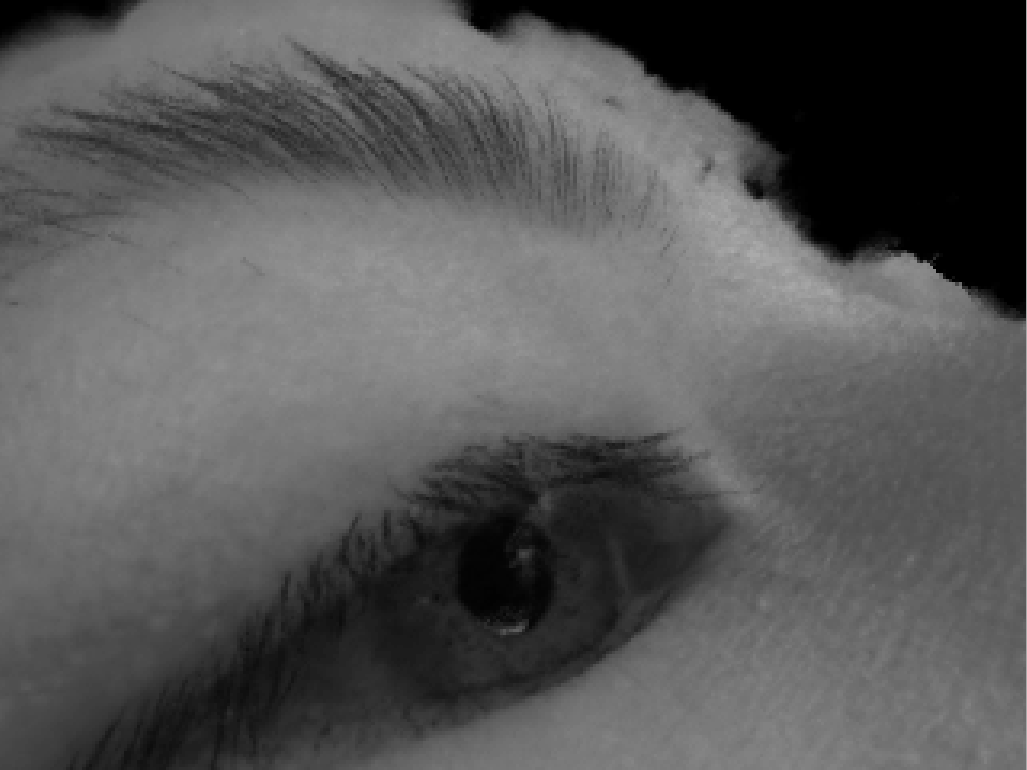} &
\includegraphics[width=\linewidth]{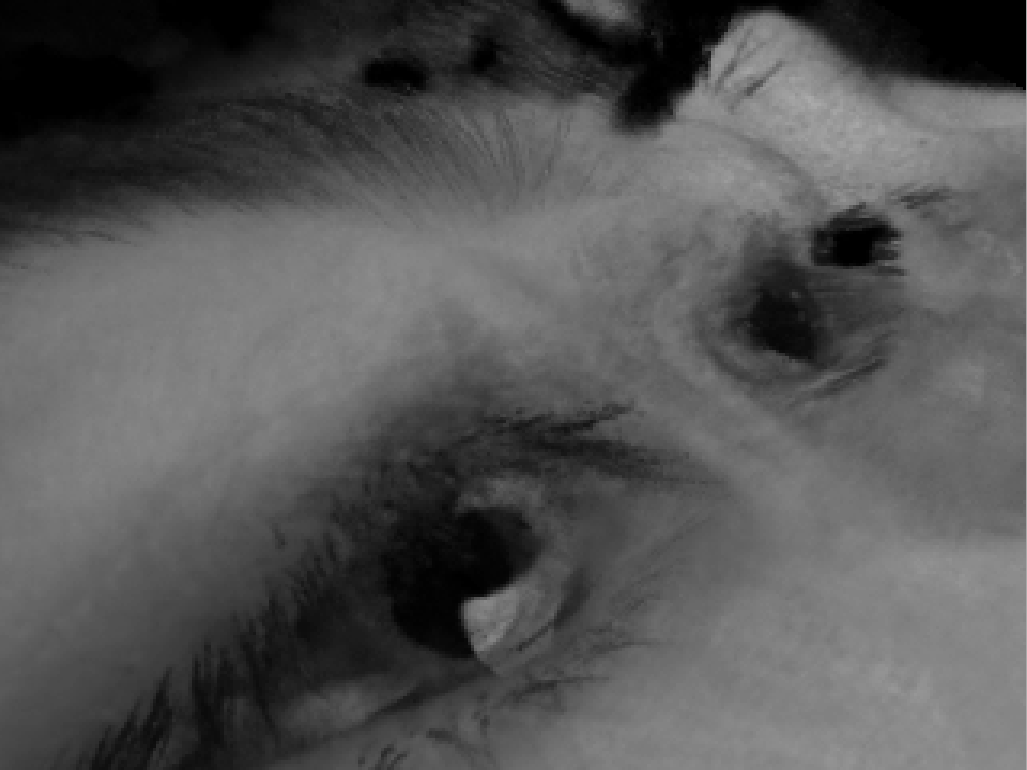} &
\includegraphics[width=\linewidth]{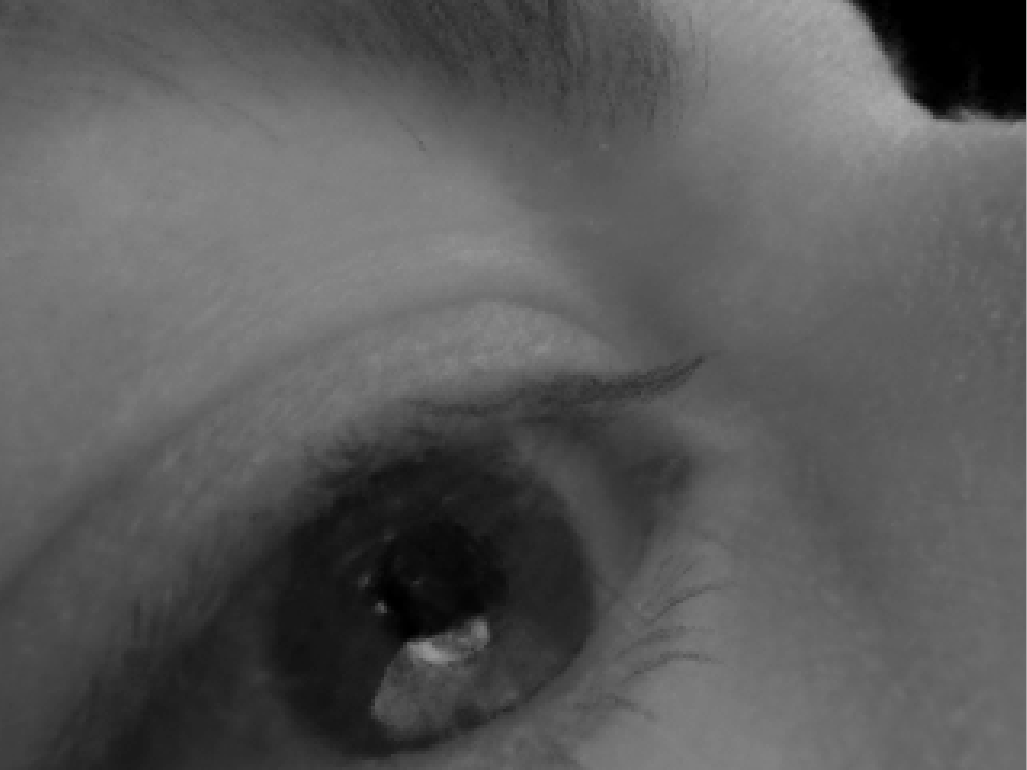} & 
\includegraphics[width=\linewidth]{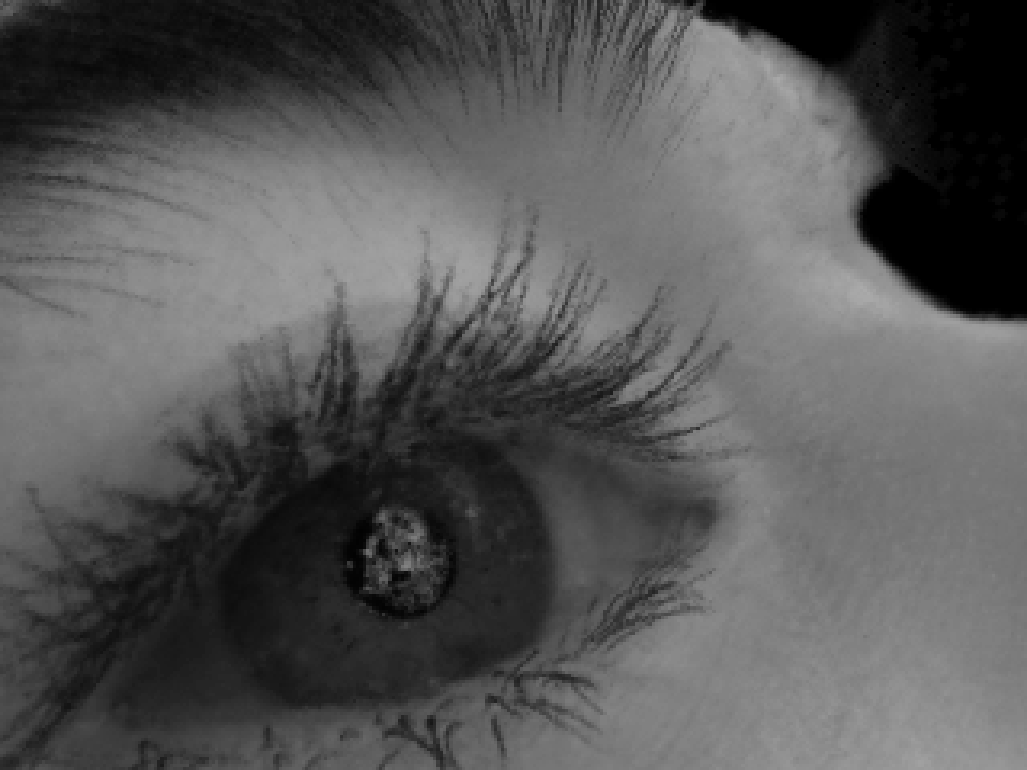} \\

\centering \footnotesize \textbf{\acron\\(Ours)} &
\includegraphics[width=\linewidth]{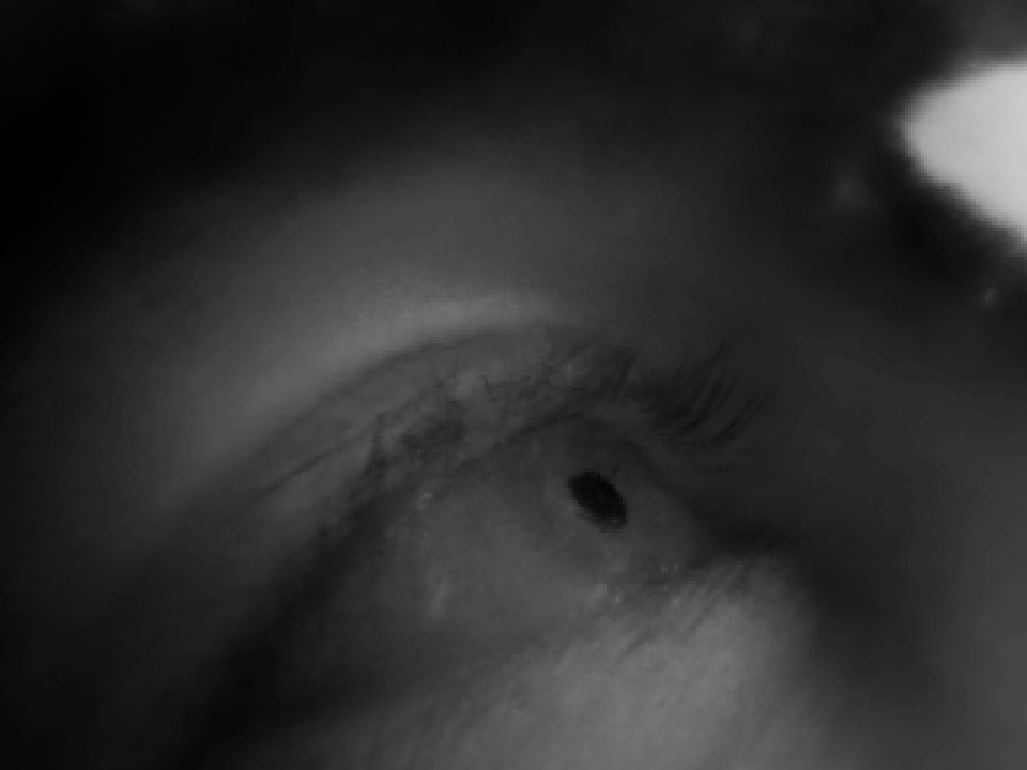} &
\includegraphics[width=\linewidth]{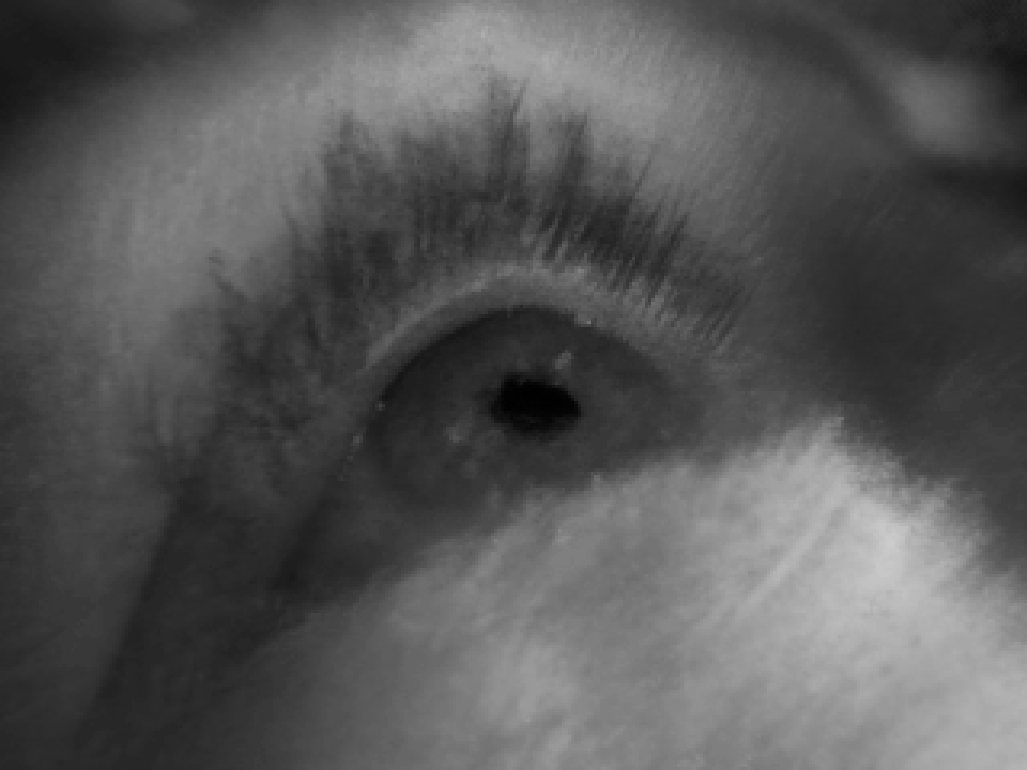} &
\includegraphics[width=\linewidth]{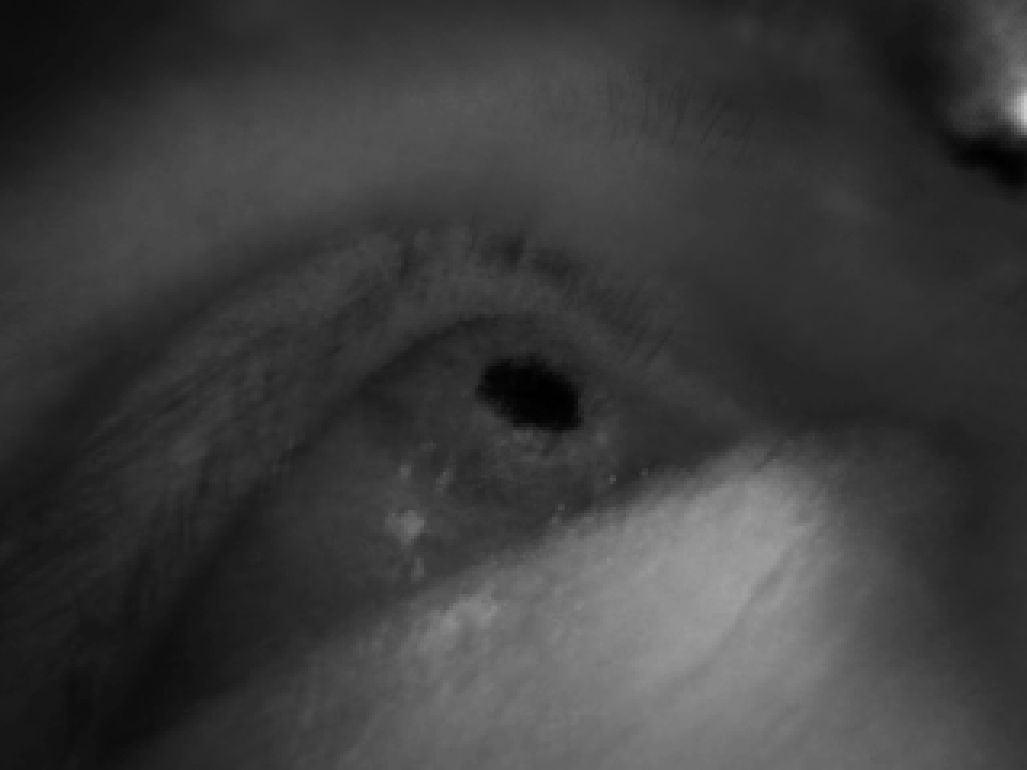} &
\includegraphics[width=\linewidth]{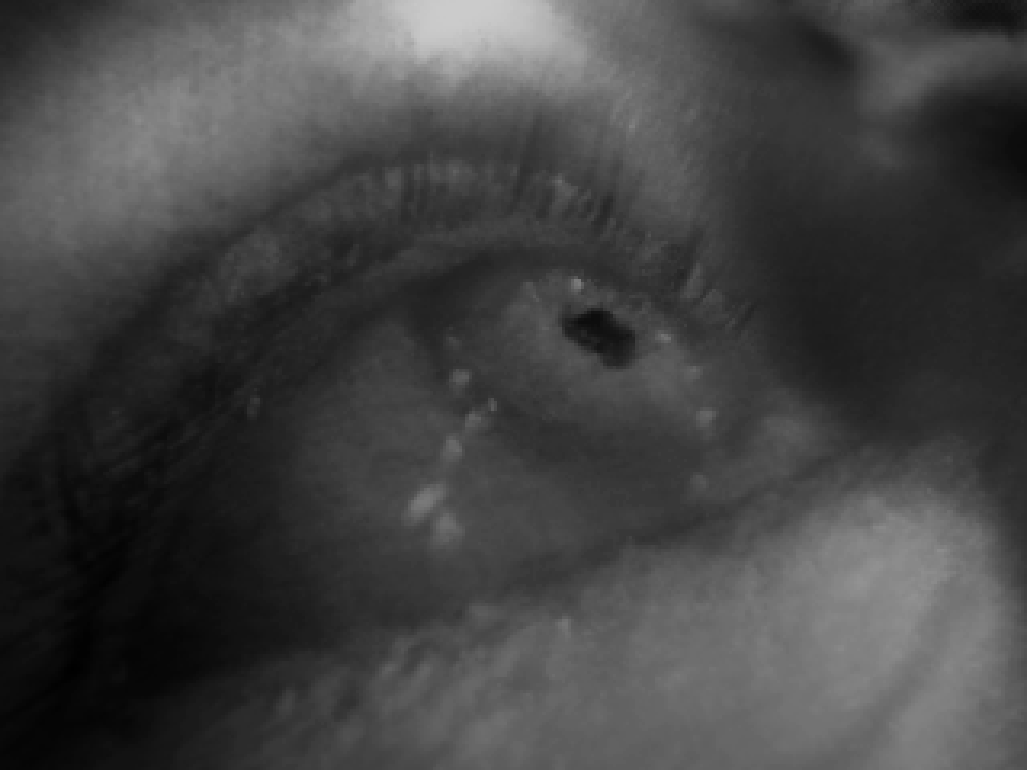} &
\includegraphics[width=\linewidth]{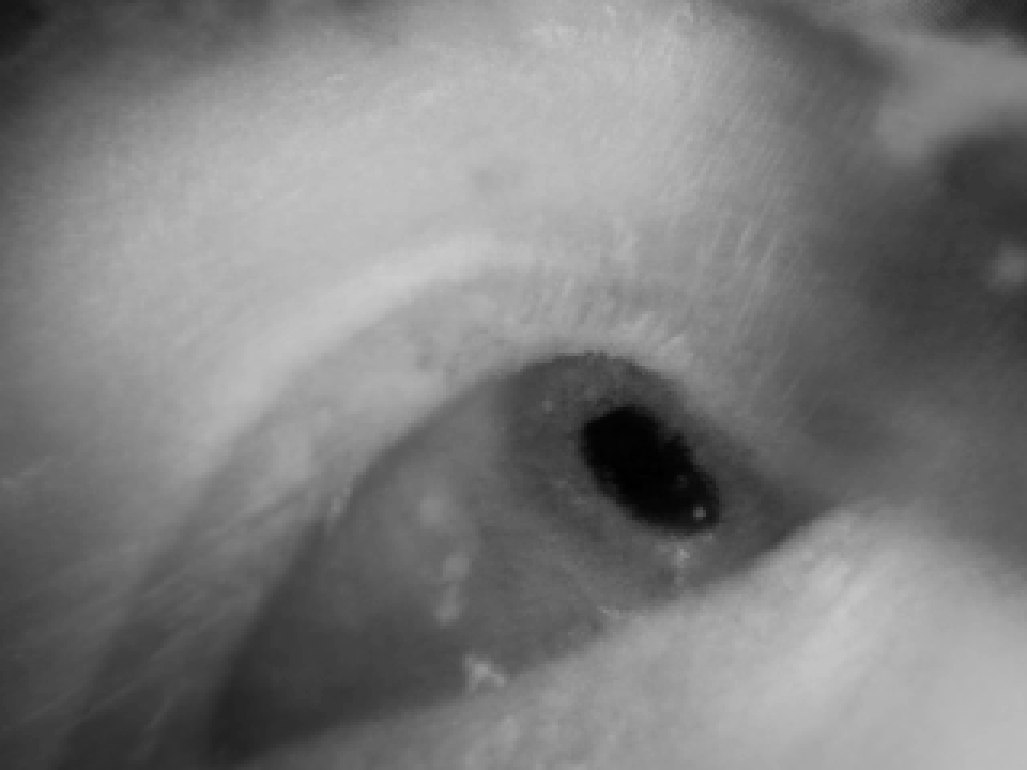} &
\includegraphics[width=\linewidth]{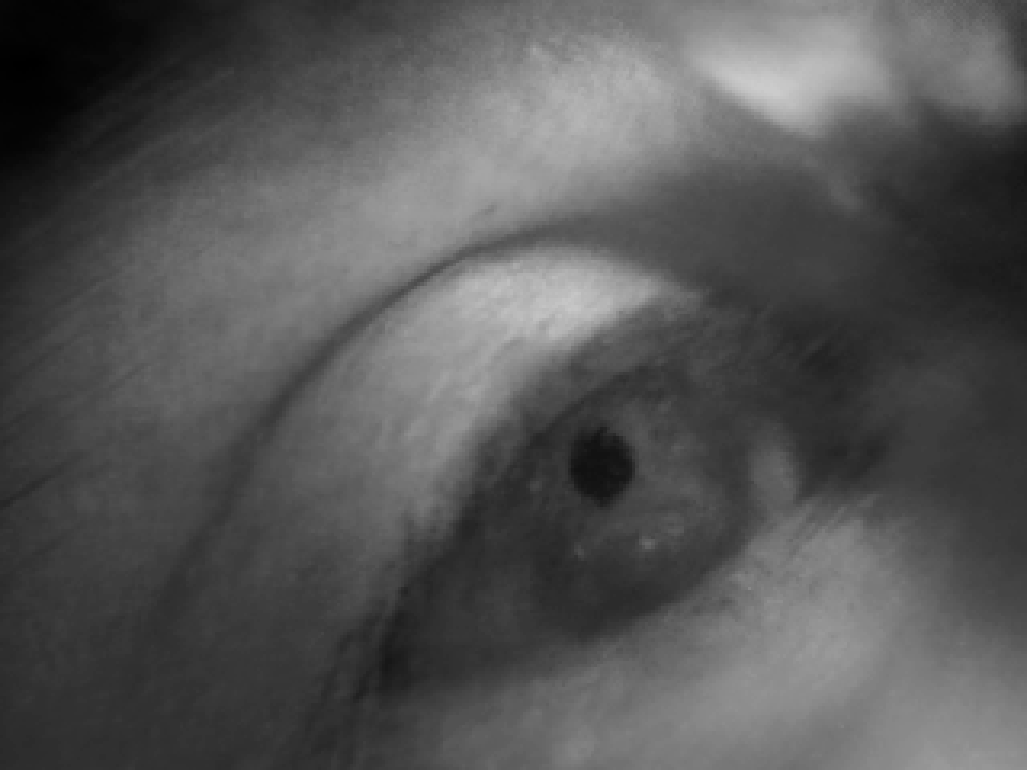} \\

\centering \footnotesize Real Aria samples~\cite{engel2023aria} &
\includegraphics[width=\linewidth]{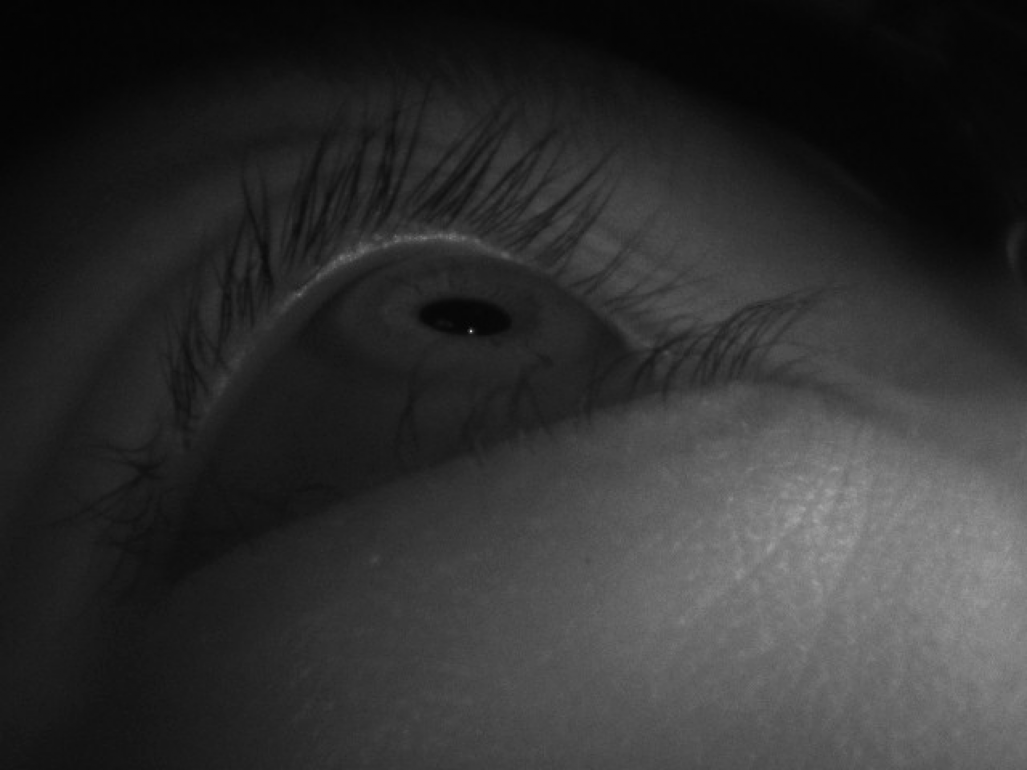} &
\includegraphics[width=\linewidth]{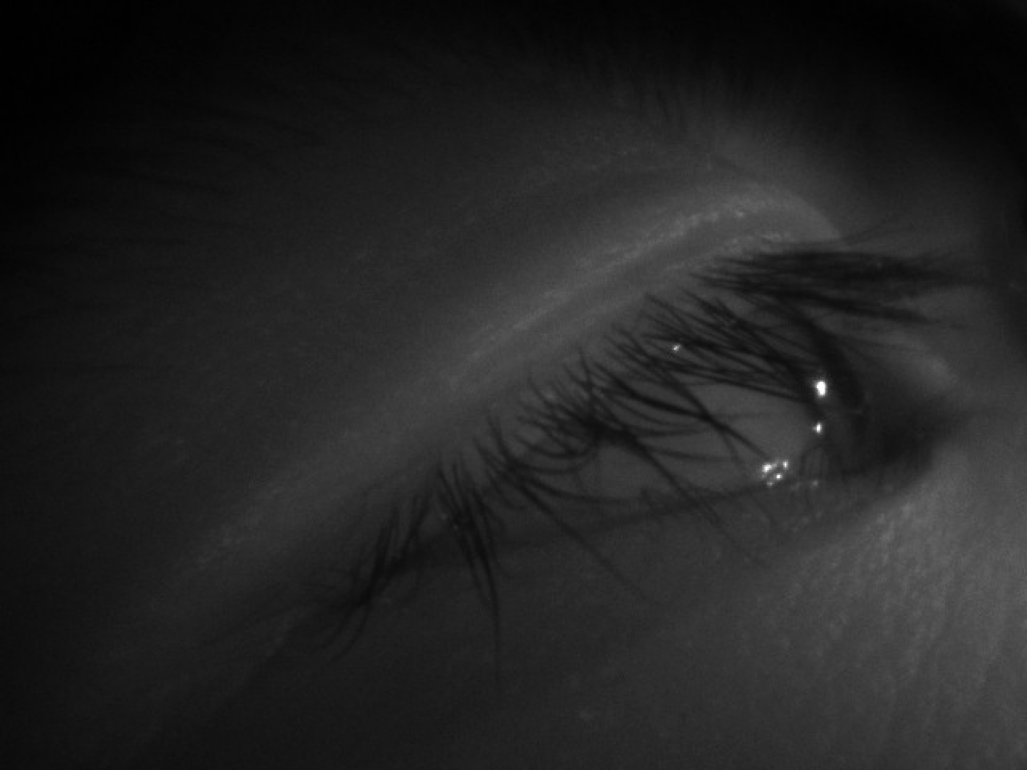} &
\includegraphics[width=\linewidth]{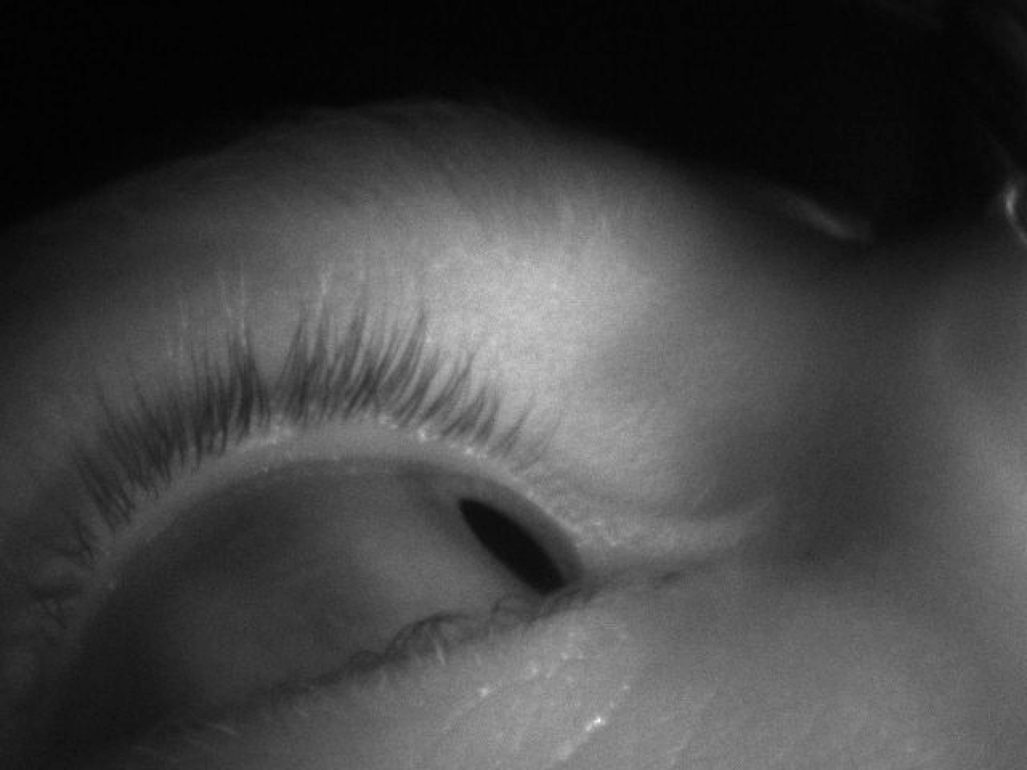} &
\includegraphics[width=\linewidth]{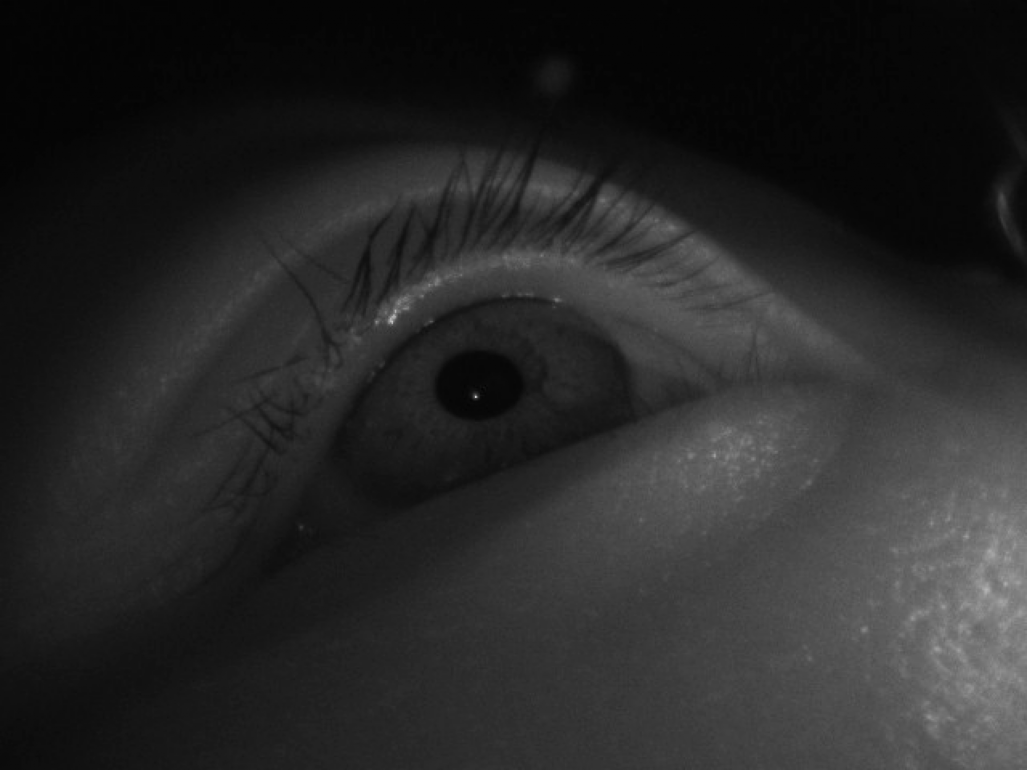} &
\includegraphics[width=\linewidth]{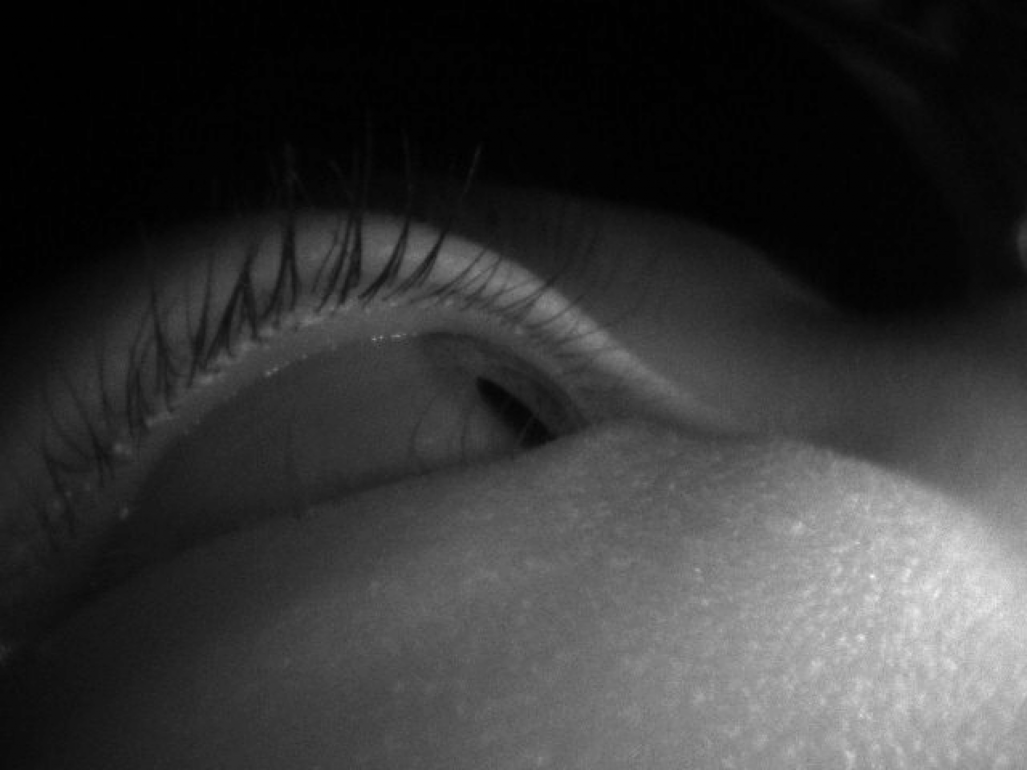} & 
\includegraphics[width=\linewidth]{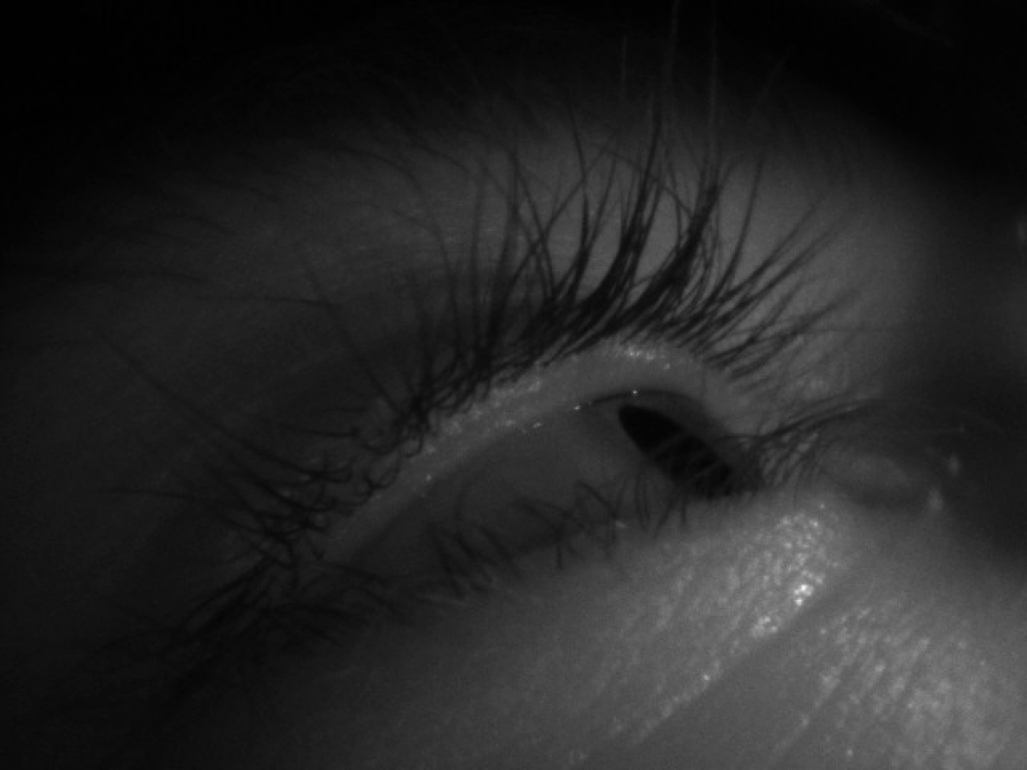} 
\end{tabular}
\vspace{-1.0em}
\caption{\textbf{Qualitative comparison.} Previous methods lack fidelity, whereas our method produces high-quality and accurate samples that match the real Aria samples~\cite{engel2023aria}.}
\label{fig:qualitative}
\end{figure*}

\subsubsection{Qualitative Evaluation.}
\label{sec:qualitative}

In \cref{fig:qualitative},  we show images synthesized by our method and by the baselines, along with real data captured by Aria \cite{engel2023aria} glasses for comparison. Note that the images by ControllableGaze~\cite{li2025we} lack fidelity. Those generated by Lin \textit{et al.}~\cite{lin2025digitally} contain floaters, lack diversity, and exhibit a significant domain gap with the real Aria images. In contrast, our method synthesizes training samples with high quality details, such as eyelashes or eyelids, and little domain gap compared to the Aria images.  This is largely because the source of our data is directly captured from similar headsets, as opposed to limited-scale studio capture in all previous works, such as \cite{lin2025digitally,li2025we,wei2025gazegaussian}. However, training these previous methods on the sparse-views present in headset capture fails, as illustrated by ~\cref{fig:view_sparsification}.

\subsection{Training an Eye Tracker}

The primary purpose of the synthesized images is to train an eye tracker. To evaluate how good they are for this purpose, we use the official Aria eye tracking repository~\cite{projectaria_eyetracking} that includes an eye tracking model. \cd{The Aria headset is widely used across research institutions and features a temporally-placed camera. It is representative of modern eye trackers and state-of-the-art research on wearables.}

We train it from  scratch using entirely synthetic data generated either by our method or by our baselines. We report our results on annotated images captured with an Aria headset in \cref{tab:et_table}. We evaluate gaze accuracy in terms of  pitch and yaw angle errors separately. We also report recall at $10^\circ$ and $5^\circ$, representing the fraction of cases where the pitch and yaw errors both lie within that range. The latter metrics are of particular importance, as they are key to unlocking new features that require a certain level of reliability. More details on the synthesized datasets and the eye tracking model can be found in the supplementary.

\begin{table}[t]
\centering
\setlength{\tabcolsep}{4pt} 
\begin{tabular}{l ccccc c}
\toprule
& \multicolumn{5}{c}{\textbf{Target domain: real Aria error \boldmath{$(^\circ)$}}} & \textbf{In-domain} \\
\cmidrule(lr){2-6} \cmidrule(lr){7-7}
Training data & Gaze $\downarrow$ & Pitch $\downarrow$ & Yaw $\downarrow$ & Rec$10^\circ \uparrow$ & Rec$5^\circ \uparrow$ & Gaze $\downarrow$ \\ 
\midrule
Lin \textit{et al.}~\cite{lin2025digitally} & 12.53 & 6.228 & 9.727 & 0.42 & 0.18 & \textit{2.39} \\
ControllableGaze~\cite{li2025we} & 12.69 & 7.806 & 9.189 & 0.44 & 0.22 & \textit{1.24} \\
\textbf{\acron~(Ours)} & \textbf{5.85} & \textbf{4.425} & \textbf{3.199} & \textbf{0.94} & \textbf{0.58} & \textit{5.02} \\
\midrule
w/o retargeting & 15.14 & 5.179 & 13.07 & 0.40 & 0.06 & \textit{6.95} \\ 
\midrule
Aria (Supervised) & 4.35 & 3.26 & 2.52 & 1.00 & 0.80 & - \\ 
\bottomrule
\end{tabular}
\vspace{0.5em}
\caption{\textbf{Eye tracking.} We synthesize eye images to simulate Aria \cite{engel2023aria} with several methods, then train the same eye tracking model on each dataset. We also report \textit{in-domain} validation where the model is evaluated on similar renders it was trained with.}
\label{tab:et_table}
\end{table}

The results highlight the effectiveness of our method. First, training directly on the data captured from a previous headset without attempting to retarget it to the Aria headset as per~\cref{sec:retargeting} yields very poor performance, highlighting the need for NVS methods such as \acron~or competing approaches~\cite{lin2025digitally,li2025we}. However, using these baselines~\cite{lin2025digitally,li2025we} to generate new training data yields only a small improvement, and the gaze angle error remains high at around $12.5^\circ$. In contrast, using \acron{} images reduces errors much more significantly across domains, with an average gaze angle error of $5.85^\circ$. 

Finally, to provide an upper bound on what can be achieved in zero-shot eye tracking, we trained the tracker with actual annotated Aria data. As could be expected the performance improves, but only by one degree in pitch and yaw. In other words, \acron{} comes very close to what can be done with ideal training data even when it is not available, which is most of the time.



\subsection{Ablation Study}

In Section~\ref{sec:quantitative}, we quantified the quality of images synthesized using our full method, after both the pretraining of Section~\ref{sec:latent} and finetuning of Section~\ref{sec:retargeting}. These can be found on the right of~\cref{tab:ablation}. On the left side of the table, we report numbers after pretraining only to gauge how good our prior model is at learning a compact human eye representation. We also provide performance numbers when we turn off key components of our approach, both after pretraining and after sparse-view finetuning.


\subsubsection{Pretraining Ablation.}

After a model is pretrained, we evaluate its ability to reconstruct samples from subjects seen during pretraining without any additional image supervision. This is challenging because the model has learned to reconstruct over 300,000 frames during the pretraining of~\cref{sec:latent} and good reconstruction performance means that the latent representation is efficient and has enough capacity for all these people and gaze directions. All models are evaluated on the same set of 20 frames. The results, in the left half of \cref{tab:ablation}, demonstrate the superiority of disentangled embeddings $(\hat{s_i}, \hat{g_i}, \hat{l_i})$ over a single per-frame latent $(f_i)$ with a sharp increase in PSNR from 18.97 dB to 25.32 dB.

\begin{figure}[t]
    \centering
    \includegraphics[width=0.98\linewidth]{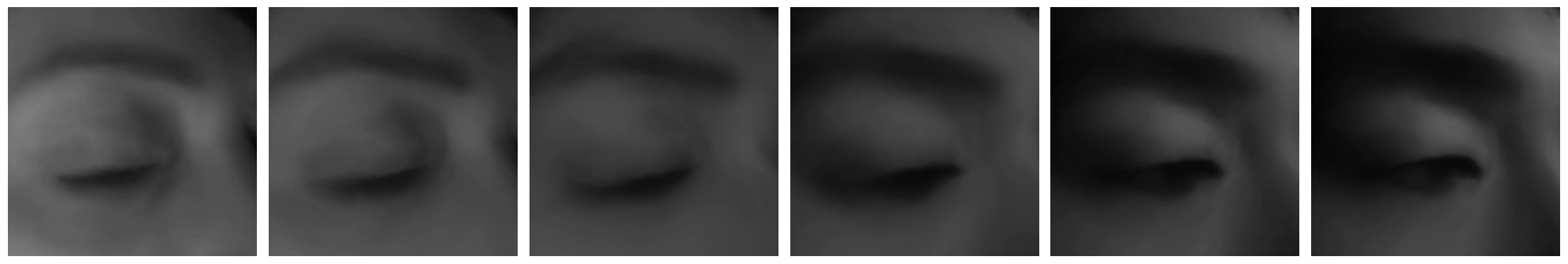}
    \caption{\textbf{Latent-space interpolation.} We interpolate latent codes across a combined manifold of subject identities, gaze directions, and lighting settings, demonstrating the prior's structural consistency while smoothly transitioning between distinct attributes.}
    \label{fig:interpolation}
\end{figure}

Since our pretrained model has learned to reconstruct a large number of subjects, gazes, and light settings, we further evaluate the learned manifold by rendering latent space interpolations. We interpolate the values of $(\hat{s_f}, \hat{g_f}, \hat{l_f})$ between two frames and visualize the associated renderings in \cref{fig:interpolation} and in our supplementary material. These interpolations highlight the smoothness of the learned manifold, as the renders seamlessly morph from one frame to another.  

\subsubsection{Finetuning Ablation.}

To evaluate the effectiveness of different pretrained models, we leave out one view out of five from our finetuning dataset for evaluation. We reconstruct a set of 20 previously unseen frames from the remaining views and report our resuls in the right-half of \cref{tab:ablation}. 
They show that disentanglement of subject, gaze and light latents is the most critical design choice in \acron{},
which may seem counter-intuitive in this context given that only the latent code for the given frame is optimized during finetuning.
However, disentangled latents yield a notable improvement from 19.09 dB to 22.10. This gain can be attributed to the superior pretraining performance of the disentangled model, which has learned a more efficient and expressive representation of the diversity in human subjects and gaze directions.
 
Also note that if we make the density  light-dependent by feeding the light latent $\hat{l_I}$ directly to the density network as opposed to only the color network, PSNR reduces to 21.53 dB. Finally, our ablation results demonstrate that sinusoidal gaze encoding is more efficient than a learned codebook, with 22.10 dB and 20.0 dB respectively. This is particularly true in finetuning, where the fixed encoding brings a strong prior to the network even if the exact gaze direction wasn't seen during pretraining.



\section{Discussion}
In this work, we introduced the first 3D gaze prior for eye image synthesis, consisting of a pretraining stage which learns a disentangled latent space of human gazes from data, and a finetuning stage which reconstructs from sparse-view inputs and renders training samples for a new eye tracking device. 

While our work represents the first step to learning a gaze prior for eye image synthesis, several possible improvements can be addressed in future work.
In some cases, eye tracking models may depend on high-frequency glints observed on the subject’s cornea to estimate gaze direction. 
These glints are challenging to capture and simulate with radiance fields. Future research could focus on advancing the representation of eye surface reflections in eye synthesis.
More broadly, we hope our work will lead to further research on zero-shot eye tracking, an underexplored task with significant applications in AR/VR that has the potential to unlock important capabilities such as foveated rendering or gaze-pinch interactions without the associated data-collection costs.

\bibliographystyle{unsrtnat}
\bibliography{paper}

\clearpage
\beginappendix

\section{Supplementary material}

\subsection{Additional latent space experiments}

\begin{figure}[h]
     \centering
     \hfill
     \begin{subfigure}[b]{0.49\textwidth}
         \centering
         \includegraphics[width=\linewidth]{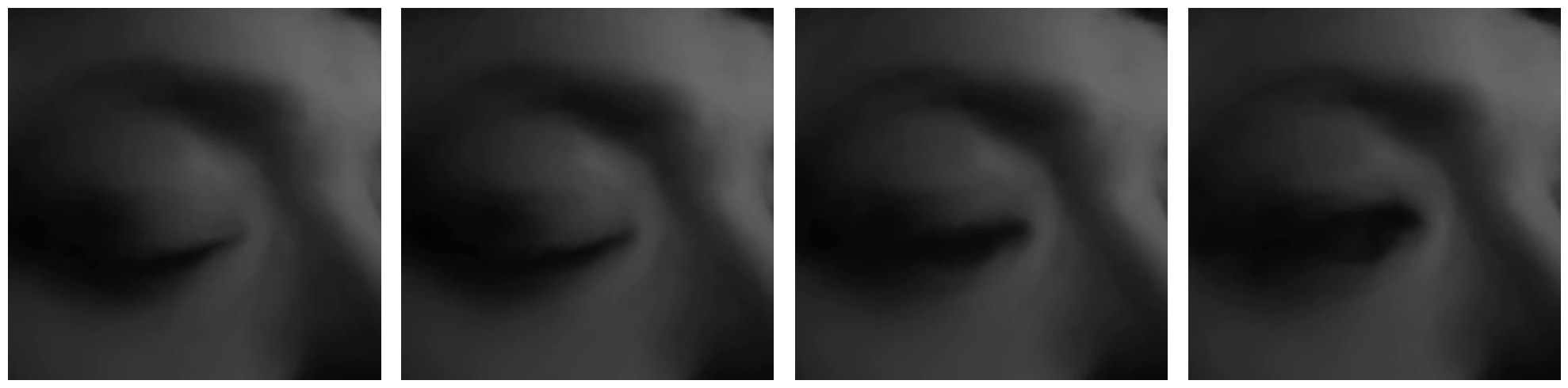}
         \caption{Gaze Interpolation}
         \label{fig:interp_gaze}
     \end{subfigure}
     \hfill
     \begin{subfigure}[b]{0.49\textwidth}
         \centering
         \includegraphics[width=\linewidth]{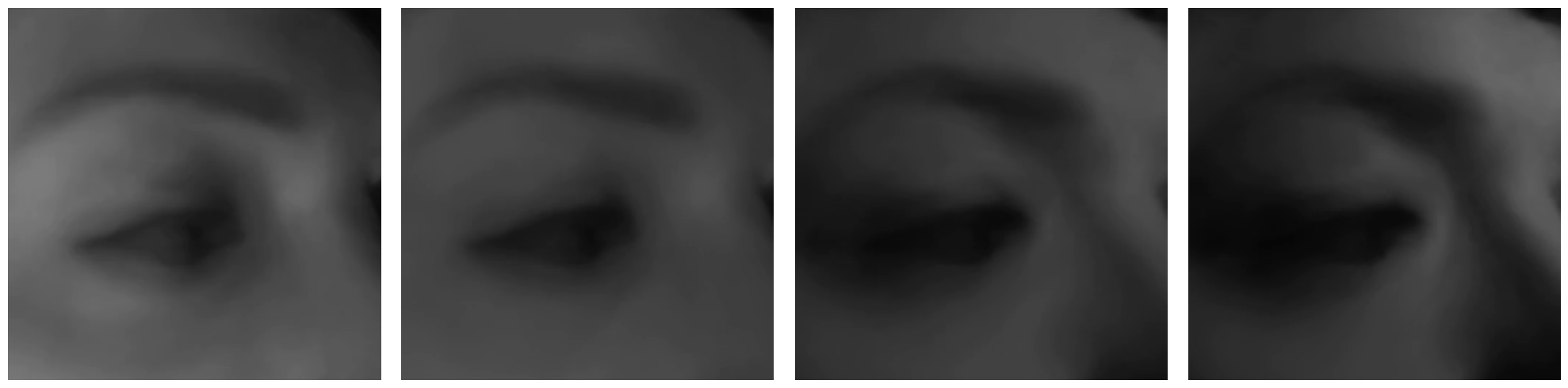}
         \caption{Light Interpolation}
         \label{fig:interp_all}
     \end{subfigure}
     \hfill


     \hfill
     \begin{subfigure}[b]{0.49\textwidth}
         \centering
         \includegraphics[width=\linewidth]{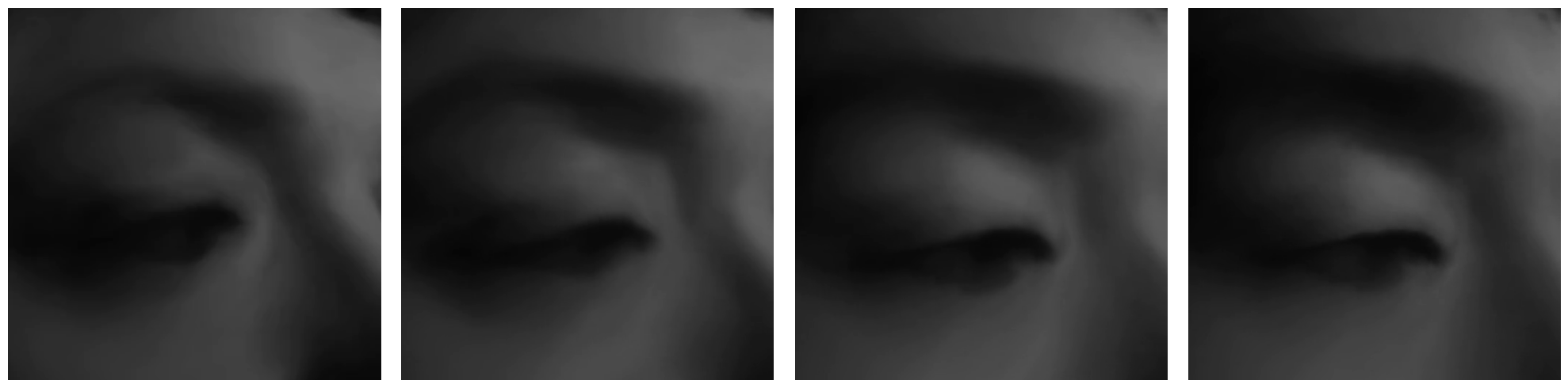}
         \caption{Subject Interpolation}
         \label{fig:interp_subject}
     \end{subfigure}
     \hfill
     \begin{subfigure}[b]{0.49\textwidth}
         \centering
         \includegraphics[width=\linewidth]{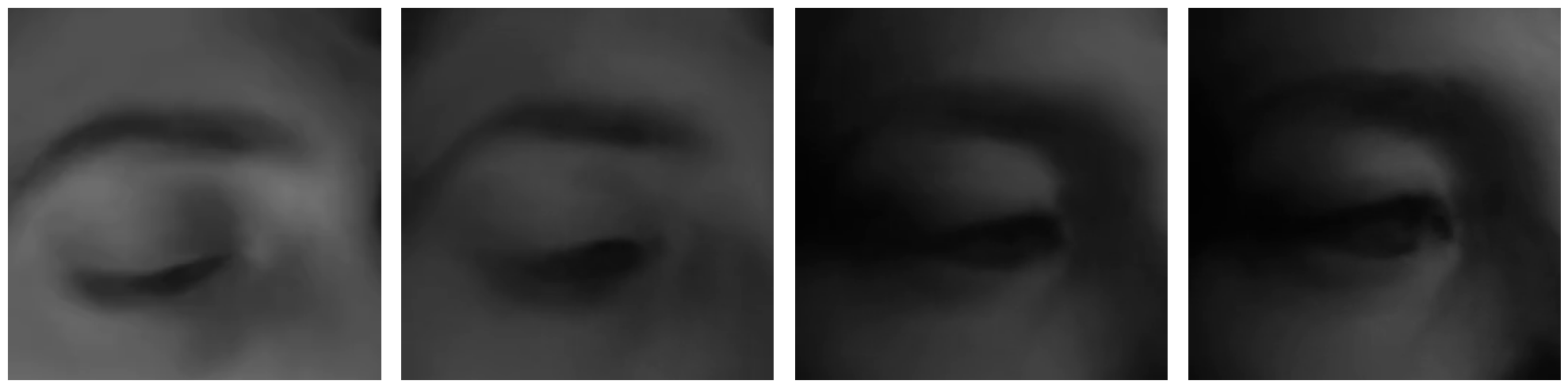}
         \caption{Gaze/Light/Subject Interpolation}
         \label{fig:interp_light}
     \end{subfigure}
     \hfill
     \caption{\textbf{Disentangled latent-space interpolation.} We render the pretraining model and interpolate latent codes independently, demonstrating its ability to learn a smooth manifold of subject identities, gaze directions, and light settings.}
     \label{fig:interpolation_grid}
\end{figure}

We provide additional interpolation visualizations in \cref{fig:interpolation_grid}. In particular, we demonstrate the ability of \acron~to independently interpolate between identities, gaze directions, and light settings. While it would be possible to train an eye tracker on these interpolated renders directly, we find that the latent optimization and finetuning stages are necessary to obtain the high-resolution results required to train a robust eye tracking model.

\subsection{Datasets}

\begin{figure}
\centering
\hfill 
\begin{minipage}[b]{0.6\columnwidth}
    \centering
    \includegraphics[width=\linewidth]{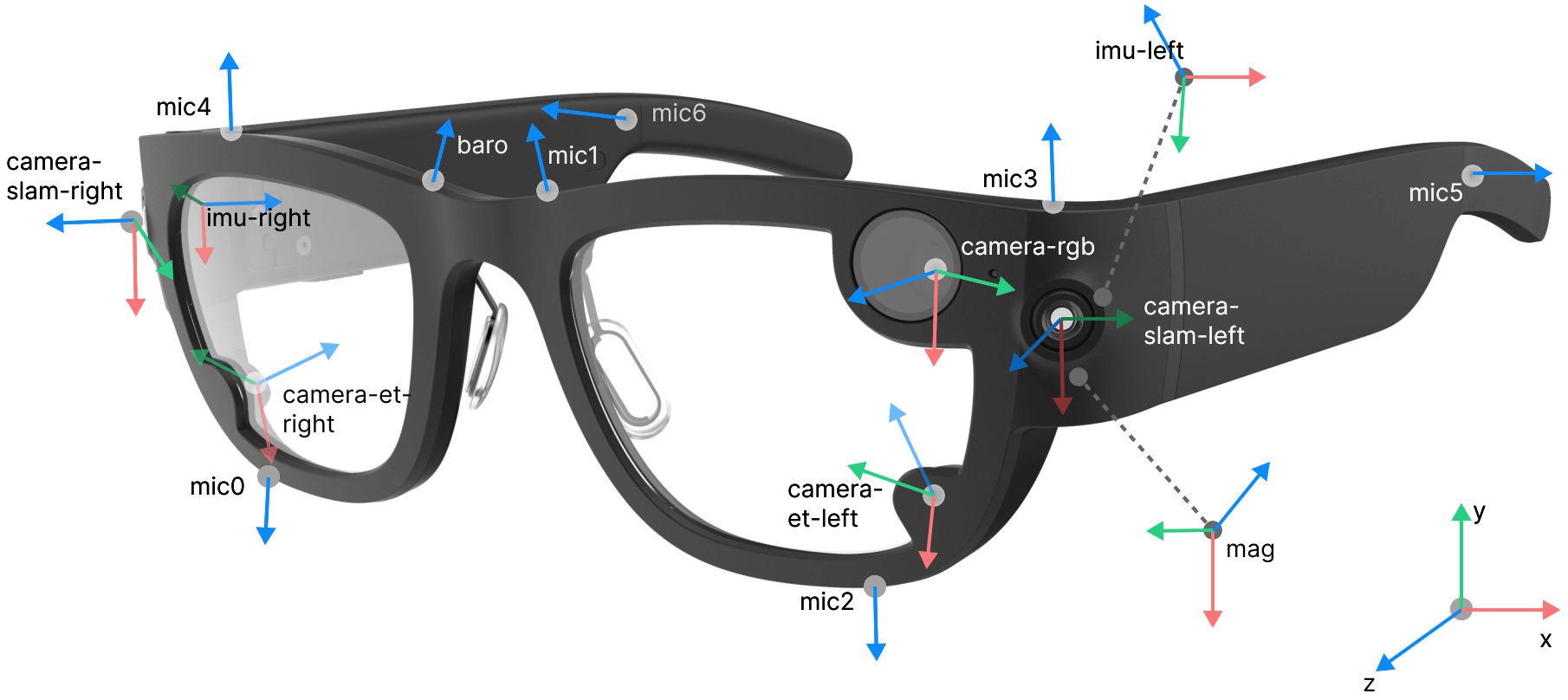} 
    \caption*{(\textbf{a}) Aria \cite{engel2023aria} coordinate systems.}
    \label{fig:subfig_a_img}
\end{minipage}
\hfill 
\begin{minipage}[b]{0.3\columnwidth}
    \centering
    \includegraphics[width=0.65\linewidth]{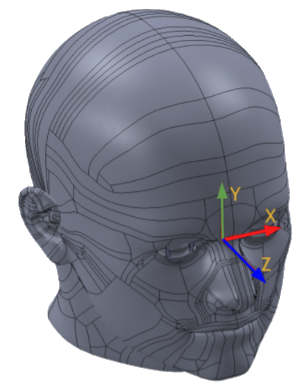} 
    \caption*{(\textbf{b}) Central pupil frame.}
    \label{fig:subfig_b_img}
\end{minipage}
\hfill 
\vspace{0.5em} 
\caption{\textbf{Coordinate systems.} The Aria~\cite{engel2023aria} project defines several reference frames. Illustrations are from the Aria project documentation~\cite{engel2023aria}.}
\label{fig:coordinates}
\end{figure}

We align all datasets in Central Pupil Frame (CPF), illustrated in \ref{fig:coordinates}. The Aria \cite{engel2023aria} documentation includes utilities to convert from device coordinates to CPF. The method of Lin \textit{et al.}~\cite{lin2025digitally} also contains utilities to provide reconstructions in CPF. For GazeGaussian~\cite{wei2025gazegaussian} and ControllableGaze~\cite{li2025we}, we compute FLAME~\cite{li2017learning} coordinates which provide a set of face landmarks. Using these, we are able to compute the 3D position of the two pupils and translate the reconstructed model to align their middle point with the 3D origin. Following this, we render the Aria camera for all reconstructed models using the camera projection in CPF coordinates.
For all datasets, we restrict our study to the right-eye, and the left-eye can easily be simulated by mirroring the right image. 

\subsubsection{Aria benchmark.} 
\begin{figure}[ht]
    \centering
    \begin{subfigure}[b]{0.23\textwidth}
        \centering
        \includegraphics[width=\textwidth]{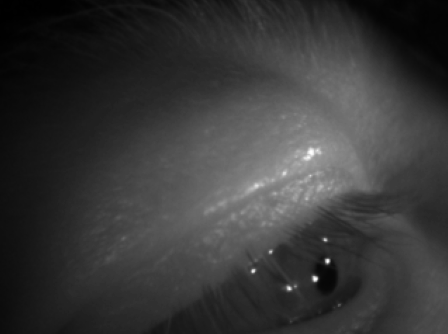}
        \caption{\centering Min pitch\\y = 10.0°, p = -52.8°}
        \label{fig:min_pitch}
    \end{subfigure}
    \begin{subfigure}[b]{0.23\textwidth}
        \centering
        \includegraphics[width=\textwidth]{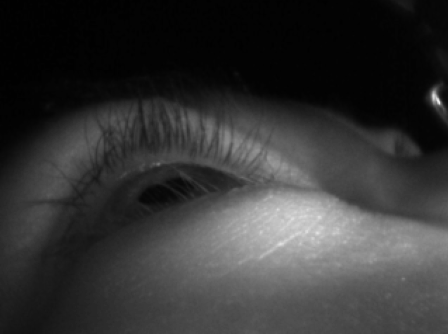}
        \caption{\centering Max pitch\\y = -27.9°, p = 38.2°}
        \label{fig:max_pitch}
    \end{subfigure}
    \begin{subfigure}[b]{0.23\textwidth}
        \centering
        \includegraphics[width=\textwidth]{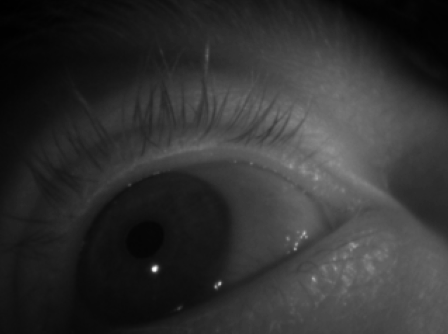}
        \caption{\centering Min yaw\\y = -50.0°, p = -4.1°}
        \label{fig:min_yaw}
    \end{subfigure}
    \begin{subfigure}[b]{0.23\textwidth}
        \centering
        \includegraphics[width=\textwidth]{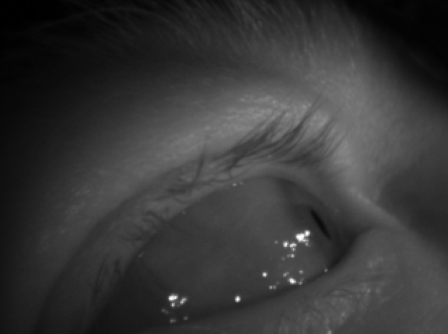}
        \caption{\centering Max yaw\\y = 52.8°, p = -28.2°}
        \label{fig:max_yaw}
    \end{subfigure}
    \caption{Visualization of extreme values in Aria~\cite{engel2023aria} data.}
    \label{fig:aria_minmax}
\end{figure}

For evaluation, we use a dataset of annotated Aria data with over 100,000 samples spanning several hundred identities. We display some of these samples in the last row of \cref{fig:qualitative}.
The majority of gaze directions in the dataset lie between plus or minus 30 degrees in pitch and yaw angles, as visualized in \cref{fig:heatmaps}, with some exceptions reaching plus or minus 50 degrees. These exceptions are visualized in~\cref{fig:aria_minmax}, where we display the minimum and maximum for both pitch and yaw gaze angles.

\subsubsection{Pretraining dataset.} 
The dataset of Lin \textit{et al.}~\cite{lin2025digitally} has 460 participants, captured with 70 different gaze fixations and 10 different light settings. We use this data for pretraining, as explained in \cref{fig:stages}. 
Lin \textit{et al.} also use this data to generate an eye synthesis dataset matching the Aria eye tracking system. We use this rendered dataset of 41180 samples to perform eye tracking experiments in \cref{sec:experiments}. We observe in \cref{fig:heatmaps} that this dataset suffers from a limited range of gaze angles, which can be attributed to the limited number of gaze fixations present in the studio capture setup, unlike our proposed retargeting method.

\subsubsection{Retargeting dataset.}
\begin{figure}[t]
\centering
\hfill 
\begin{minipage}[b]{0.22\columnwidth}
    \centering
    \includegraphics[width=\linewidth]{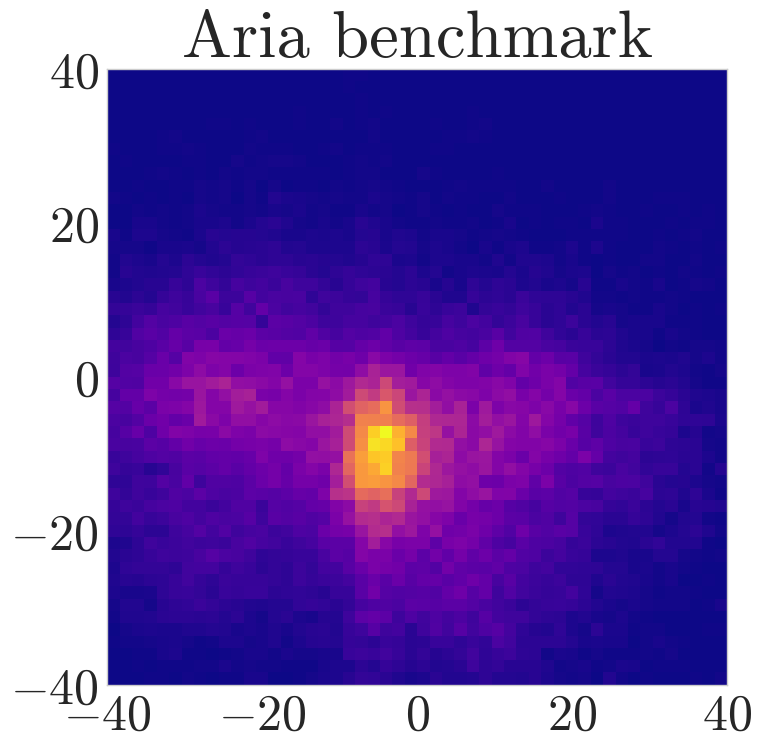} 
    \label{fig:heat_aria}
\end{minipage}
\hfill 
\begin{minipage}[b]{0.22\columnwidth}
    \centering
    \includegraphics[width=\linewidth]{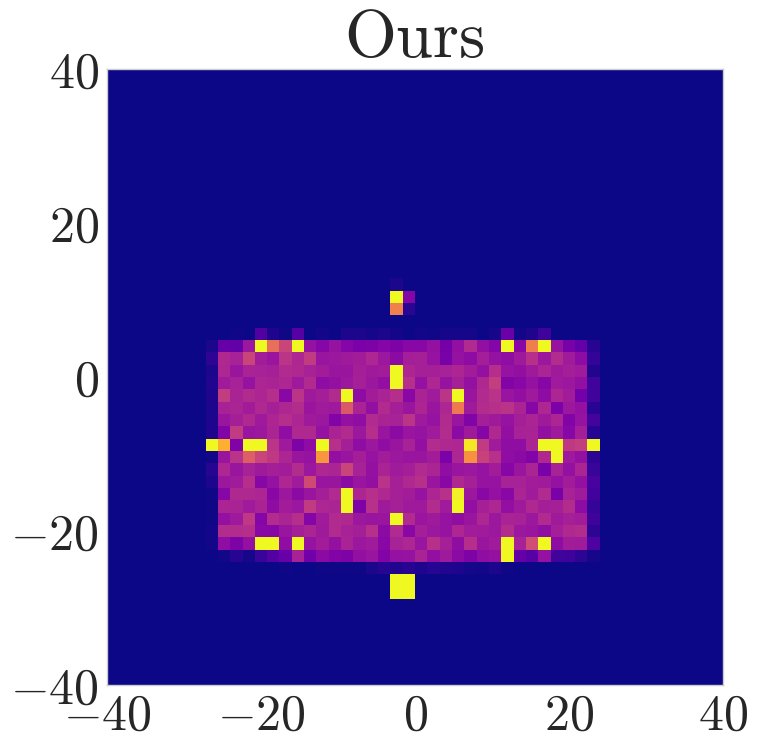} 
    \label{fig:heat_masq}
\end{minipage}
\hfill 
\begin{minipage}[b]{0.22\columnwidth}
    \centering
    \includegraphics[width=\linewidth]{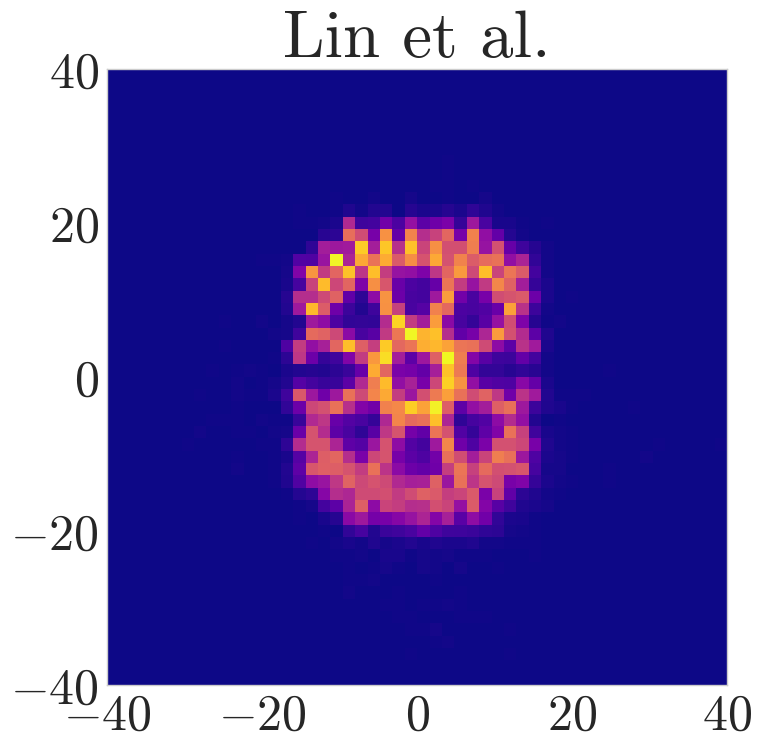} 
    \label{fig:heat_ours}
\end{minipage}
\hfill 
\begin{minipage}[b]{0.22\columnwidth}
    \centering
    \includegraphics[width=\linewidth]{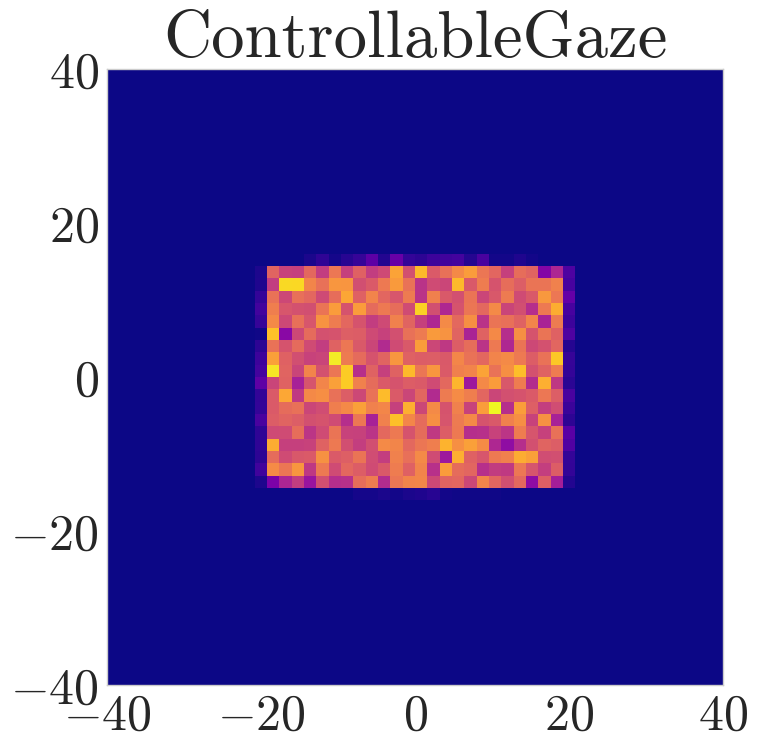} 
    \label{fig:heat_controllablegaze}
\end{minipage}
\hfill 
\begin{minipage}[b]{0.028\columnwidth}
    \centering
    \raisebox{0.2\height}{\includegraphics[width=\linewidth]{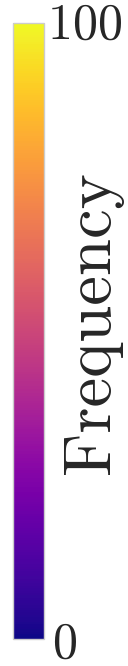}}
    \label{fig:colorbar}
\end{minipage}
\hfill 
\caption{Heatmaps of right-eye pitch-yaw pairs found in different datasets.}
\label{fig:heatmaps}
\end{figure}

\begin{figure}[ht]
    \centering
    \includegraphics[width=0.95\textwidth]{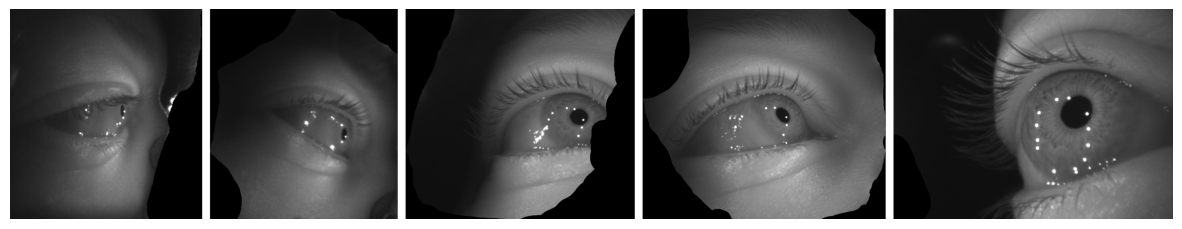}
    \vspace{0.em}
    \includegraphics[width=0.95\textwidth]{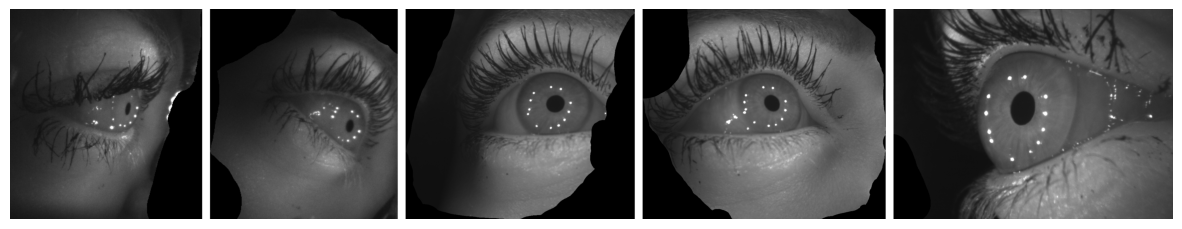}
    \vspace{0.em}
    \includegraphics[width=0.95\textwidth]{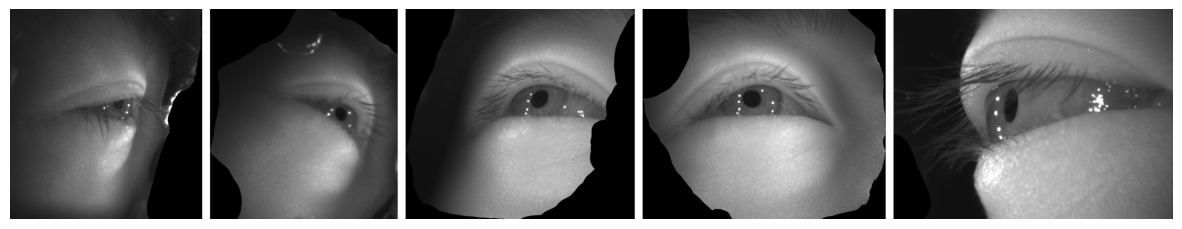}
    \caption{Samples from the source dataset before retargeting.}
    \label{fig:viz_source}
\end{figure}

Our sparse-view source data, which we use to generate our retargeted dataset, includes 58,195 samples with a wide range of gaze directions and from 14,708 different participants. We visualize some samples from this dataset in \cref{fig:viz_source}. The ground-truth gaze directions are obtained in two ways. In the first scenario, the user looks at a fixation from a finite set. In the second scenario, a random point is lit on a monitor and the user is asked to look at the point. This leads to the data distribution shown in \cref{fig:heatmaps}.

\subsubsection{ControllableGaze dataset.}
ControllableGaze \cite{li2025we} uses MPIIGaze \cite{zhang2017mpiigaze}, a dataset spanning 213,659 images collected from 15 participants in a studio setup with 1498 - 34745 images per participant. We adapt their code to render their models using the Aria camera, and follow their procedure to generate 3000 images per subject with varying pitch and yaw angles, or 45,000 training samples in total. We compute FLAME~\cite{li2017learning} face landmarks on the 3D model to set the 3D origin to the middle of the two pupils, thus aligning the generated data with the Aria camera matrix. Finally, since the initial images are captured in RGB, we convert the synthesized images to grayscale to match the appearance of AR/VR infrared cameras. We align eye models reconstructed with ControllableGaze~\cite{li2025we} in CPF by translating the middle pupil estimated by FLAME~\cite{li2017learning} to match the 3D origin.

\subsection{Pretraining implementation details}

Our learned prior model takes inspiration from neural radiance fields~\cite{mildenhall2021nerf}, which learn a function
$f(\mathbf{x}, \mathbf{d})$ that maps a 3D position $\mathbf{x}$ and viewing direction $\mathbf{d}$ to a volume density $\sigma$ and color $\mathbf{c}$. The final color $C(\mathbf{r})$ observed along a camera ray $\mathbf{r}(t) = \mathbf{o} + t\mathbf{d}$ is computed via volume rendering, where $\mathbf{o}$ and $\mathbf{d}$ are the ray origin and direction, respectively, and $t_n$ and $t_f$ define the integration bounds:

\begin{equation*}
C(\mathbf{r}) = \int_{t_n}^{t_f} T(t) \, \sigma(\mathbf{r}(t)) \, \mathbf{c}(\mathbf{r}(t), \mathbf{d}) \, dt, \quad \text{with} \quad T(t) = \exp\left(-\int_{t_n}^{t} \sigma(\mathbf{r}(s)) ds\right).
\end{equation*}

where $T(t)$ represents the accumulated transmittance. More precisely, our pretraining network architecture is based on Instant-NGP~\cite{muller2022instant} with a multi-resolution hash-grid encoder. The hash grid comprises 16 levels of 2-dimensional features, with a base resolution of 16, and a finest resolution of 1024.  We progressively activate finer hash-grid resolutions throughout pretraining, activating grids of resolutions up to 256 initially, then 512 and 1024 at milestones of 100{,}000 and 300{,}000 iterations.
Both the sigma and color branches consist of 4 layers with a hidden dimension of 384, and geometric features of dimension 31 are passed from the density network to the color network. The subject, gaze, and light latent code dimensions are set to 256, 16, and 8, respectively.

During pretraining, we load 16 frames at a time and reload a new batch every 1000 iterations. For each iteration, we sample a chunk of 2048 rays across frames and compute the image loss $\mathcal{L}_{\text{rgb}} = |c - \hat{c}|$, where $c$ is the ground truth pixel color and $\hat{c}$ is the predicted color. Pixels with saturation above 0.85 are masked, and the background color is set to 0. The loss function also includes a KL divergence term to encourage the latent codes to follow a normal distribution, weighted by $\lambda_{\text{KL}} = 10^{-8}$. We use a perceptual loss \cite{zhang2018unreasonable} with a weight of $0.002$. Finally, the color network is regularized with a weight of $1.0 \times 10^{-5}$. This pretraining is performed for 1 million iterations with a learning rate of 0.001 and a learning rate decay of 250.  

We also experimented with gradient accumulation, in an attempt to average more gradients over a wider diversity of subjects during pretraining. However, we found this to be detrimental, and we opt for more frequent small gradient steps instead.

\subsection{Retargeting implementation details}
For retargeting, we employ a two-stage optimization, starting with latent optimization and continuing with full finetuning, to maximize the inductive bias of our pretrained 3D prior. In each iteration of latent optimization, 2048 rays are sampled across all available source views, and the image reconstruction loss $\mathcal{L}_{\text{rgb}}$ is minimized. We find that 50 optimization steps are sufficient to reach a satisfactory initialization.
Once the starting latent codes are computed, we unfreeze the model and jointly optimize the weights of $\truefunc_{\Theta, I}$ and the latent codes. The same loss as pretraining $\mathcal{L}$ is used, except we disable the KL loss and progressive hash-grid activation. Finetuning is performed for to 18{,}000 iterations with a learning rate of 0.004. 

We apply a predefined validity mask on the source images to remove the headset when it is visible on any of the frames. These masks are generated only once, then applied across the entire dataset, since the cameras are in the same position relative to the glass frame and its visibility is thus identical.
We also observe that correct modeling of the distortion coefficients is essential to reconstruct and render realistic images at this stage.

\subsection{Eye tracking implementation details}
\label{sec:eye_tracking_impl}

In our eye tracking experiments, we use the public Aria eye tracking model~\cite{projectaria_eyetracking}, which employs a modified ResNet-18 architecture. The main backbone consists of a convolutional block followed by four residual stages. The convolutional block comprises a $7 \times 7$ convolution with stride 2, batch normalization, ReLU activation, and $3 \times 3$ max pooling with stride 2. The four residual stages have channel dimensions $[32, 64, 128, 128]$ with block counts $[1, 2, 1, 2]$ and strides $[1, 2, 2, 2]$, respectively. Then, the prediction head applies adaptive average pooling to reduce spatial dimensions.
The final model has approximately 2.3M parameters with input resolution $240 \times 320$ pixels, and outputs a two-dimensional gaze-direction.

\begin{figure}[htbp]
    \centering
    \begin{subfigure}[b]{0.16\textwidth}
        \centering
        \includegraphics[width=\textwidth]{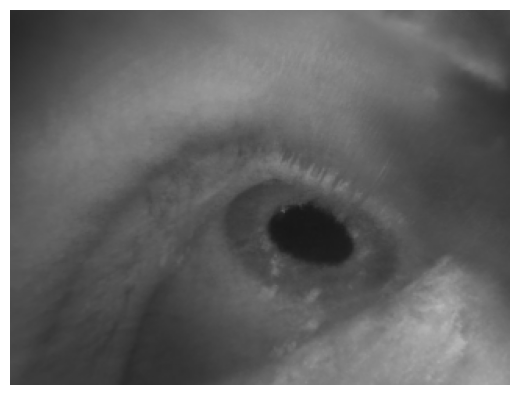}
        \caption*{Original}
        \label{fig:orig}
    \end{subfigure}
    \begin{subfigure}[b]{0.16\textwidth}
        \centering
        \includegraphics[width=\textwidth]{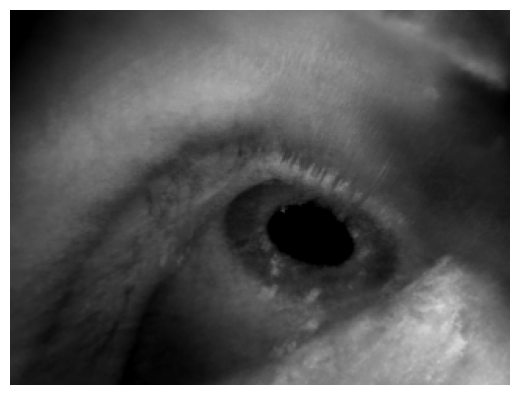}
        \caption*{Jitter}
        \label{fig:jitter}
    \end{subfigure}
    \begin{subfigure}[b]{0.16\textwidth}
        \centering
        \includegraphics[width=\textwidth]{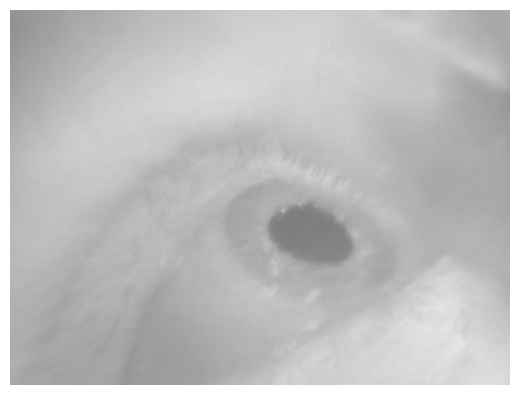}
        \caption*{Low Gamma}
        \label{fig:low_gamma}
    \end{subfigure}
    \begin{subfigure}[b]{0.16\textwidth}
        \centering
        \includegraphics[width=\textwidth]{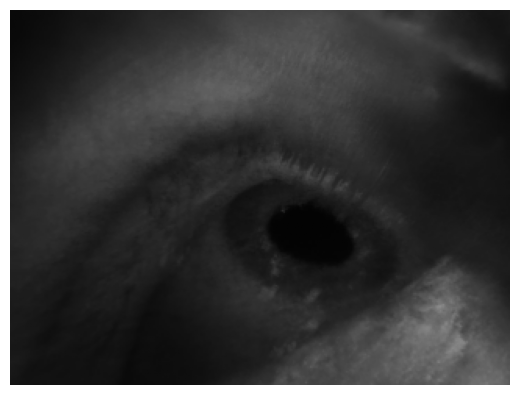}
        \caption*{High Gamma}
        \label{fig:high_gamma}
    \end{subfigure}
    \begin{subfigure}[b]{0.16\textwidth}
        \centering
        \includegraphics[width=\textwidth]{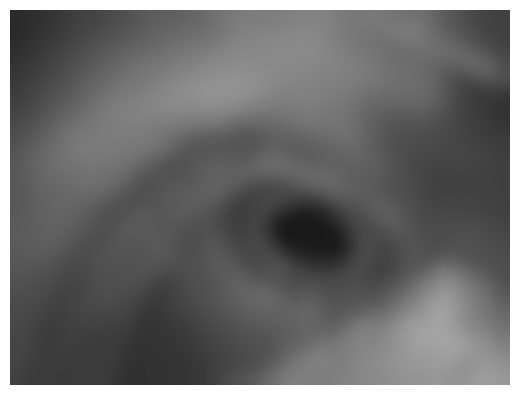}
        \caption*{Blur}
        \label{fig:blur}
    \end{subfigure}
    \begin{subfigure}[b]{0.16\textwidth}
        \centering
        \includegraphics[width=\textwidth]{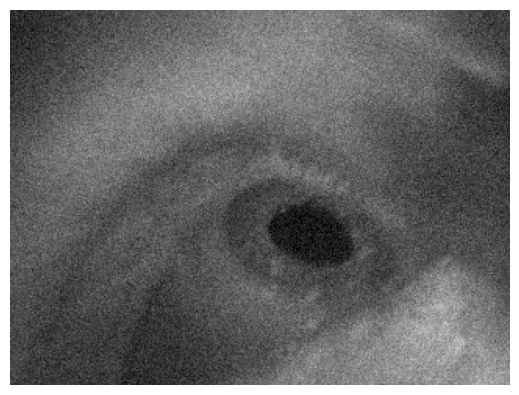}
        \caption*{Noise}
        \label{fig:gaussian}
    \end{subfigure}
    \caption{Image augmentations during eye tracking.}
    \label{fig:augmentations}
\end{figure}

We employ a Smooth-L1 loss to supervise the predicted gaze angles. We use the AdamW optimizer with learning rate $\eta = 10^{-4}$, weight decay $\lambda = 0.01$, and gradient clipping with maximum norm 1.0. Training proceeds for 500,000 steps with batch size 64. We employ automatic mixed precision (AMP) training for computational efficiency.
During training, we randomly apply color jitter, gamma randomization, gaussian blur and gaussian noise to the training images, as illustrated in~\cref{fig:augmentations} and normalize them to the range $[0, 1]$.

Since different datasets are not all equally diverse and lead to overfitting at different stages of training, we report the results of the best checkpoint for each method. Finally, we also report full in-domain metrics in \cref{tab:full_et_table}, which were not included in our main manuscript due to space limitations.

\setlength{\tabcolsep}{4pt}
\begin{table}[t]
\centering
\begin{tabular}{lccccc}
\toprule
Method & Gaze ($^\circ$) $\downarrow$ & Pitch ($^\circ$) $\downarrow$ & Yaw ($^\circ$) $\downarrow$ & Rec $10^\circ$ $\uparrow$ & Rec $5^\circ$ $\uparrow$ \\ \midrule
w/o retargeting & 15.14 & 5.179 & 13.07 & 0.40 & 0.06 \\ 
~-~ \textit{ in domain} & \textit{6.955} & \textit{3.631} & \textit{5.284} & \textit{0.78} & \textit{0.52} \\
\midrule
Lin \textit{et al.}~\cite{lin2025digitally} & 12.53 & 6.228 & 9.727 & 0.42 & 0.18 \\
~-~ \textit{ in domain} & \textit{2.389} & \textit{1.660} & \textit{1.355} & \textit{0.99} & \textit{0.96} \\
ControllableGaze~\cite{li2025we} & 12.69 & 7.806 & 9.189 & 0.44 & 0.22 \\
~-~ \textit{ in domain} & \textit{1.239} & \textit{0.681} & \textit{0.883} & \textit{1.00} & \textit{0.99} \\
\textbf{\acron~(Ours)} & \textbf{5.85} & \textbf{4.425} & \textbf{3.199} & \textbf{0.94} & \textbf{0.58} \\
~-~ \textit{ in domain} & \textit{5.021} & \textit{2.913} & \textit{3.078} & \textit{0.94} & \textit{0.68} \\
\midrule
Aria (Supervised) & 4.347 & 3.259 & 2.516 & 1.00 & 0.80 \\ 
\bottomrule
\end{tabular}
\vspace{0.5em}
\caption{\textbf{Eye tracking performance with in-domain metrics.} We report the full in-domain error metrics of the different methods evaluated.}\label{tab:full_et_table}
\end{table}

\subsection{Metrics}

To evaluate the quality of predicted pixel colors $\hat{c}$ against ground truth colors $c$, we use three standard metrics: Mean Squared Error (MSE), Peak Signal-to-Noise Ratio (PSNR), and Structural Similarity Index (SSIM).

First, MSE measures the average squared difference between the predicted and ground truth pixel colors:
\[
\mathrm{MSE} = \frac{1}{N} \sum_{i=1}^{N} \left\| c_i - \hat{c}_i \right\|^2
\]
where $N$ is the number of pixels, $c_i$ is the ground truth color, and $\hat{c}_i$ is the predicted color for pixel $i$. Lower MSE indicates better reconstruction quality.

PSNR is a logarithmic metric derived from MSE:
\[
\mathrm{PSNR} = 10 \cdot \log_{10} \left( \frac{L^2}{\mathrm{MSE}} \right)
\]
where $L$ is the maximum possible pixel value (e.g., $L=1$ for normalized images or $L=255$ for 8-bit images). Higher PSNR values indicate better image quality.

SSIM measures the perceptual similarity between two images, considering luminance, contrast, and structure:
\[
\mathrm{SSIM}(c, \hat{c}) = \frac{(2\mu_c\mu_{\hat{c}} + C_1)(2\sigma_{c\hat{c}} + C_2)}{(\mu_c^2 + \mu_{\hat{c}}^2 + C_1)(\sigma_c^2 + \sigma_{\hat{c}}^2 + C_2)}
\]
where $\mu_c$ and $\mu_{\hat{c}}$ are the mean pixel values, $\sigma_c^2$ and $\sigma_{\hat{c}}^2$ are the variances, $\sigma_{c\hat{c}}$ is the covariance, and $C_1$, $C_2$ are small constants to stabilize the division. SSIM values range from $-1$ to $1$, with higher values indicating greater similarity.

\end{document}